\documentclass[%
 reprint,
 amsmath,amssymb,
 aps,
]{revtex4-2}

\usepackage{graphicx}
\usepackage{dcolumn}
\usepackage{bm}

\renewcommand{\v}[1]{\ensuremath{\mathbf{#1}}} 
 
\newcommand{\abs}[1]{\left| #1 \right|} 
 
\renewcommand\eqref[1]{Eq.\;\ref{#1}} 

\usepackage [english]{babel}
\usepackage [english = american]{csquotes}
\MakeOuterQuote{"}

\usepackage{color}					

\begin{document}

\title{Recurrences reveal shared causal drivers of complex time series}

\author{William Gilpin}
 \email{wgilpin@utexas.edu}
\affiliation{%
Department of Physics, The University of Texas at Austin, Austin, Texas 78712, USA
}%
\affiliation{%
Oden Institute for Computational Engineering and Sciences, The University of Texas at Austin, Austin, Texas 78712, USA
}%
\affiliation{%
Marble Therapeutics, Cambridge, Massachusetts 02115, USA
}%

\date{\today}

\begin{abstract}
Unmeasured causal forces influence diverse experimental time series, such as the transcription factors that regulate genes, or the descending neurons that steer motor circuits. 
Combining the theory of skew-product dynamical systems with topological data analysis, we show that simultaneous recurrence events across multiple time series reveal the structure of their shared unobserved driving signal.
We introduce a physics-based unsupervised learning algorithm that reconstructs causal drivers by iteratively building a recurrence graph with glass-like structure. 
As the amount of data increases, a percolation transition on this graph leads to weak ergodicity breaking for random walks---revealing the shared driver's dynamics, even from strongly-corrupted measurements. 
We relate reconstruction accuracy to the rate of information transfer from a chaotic driver to the response systems, and we find that effective reconstruction proceeds through gradual approximation of the driver's dynamical attractor. Through extensive benchmarks against classical signal processing and machine learning techniques, we demonstrate our method's ability to extract causal drivers from diverse experimental datasets spanning ecology, genomics, fluid dynamics, and physiology.
\end{abstract}

                          
\maketitle

\section{Introduction}

\begin{figure*}
{
\centering
\includegraphics[width=\linewidth]{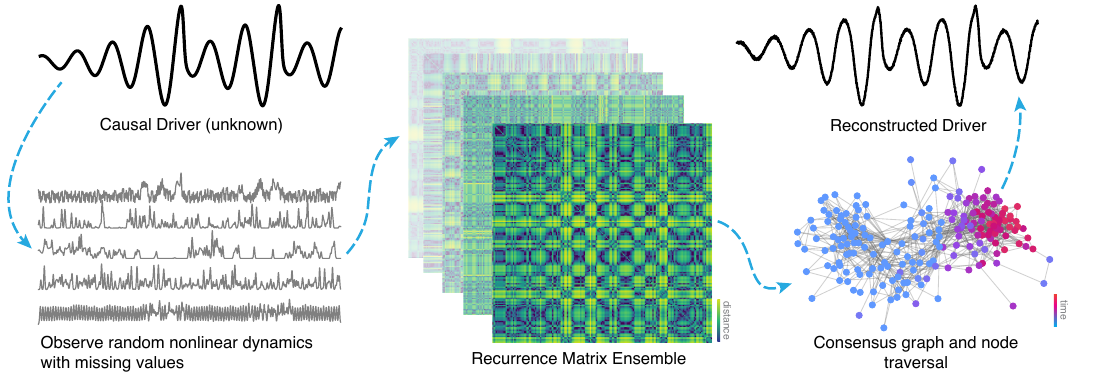}
\caption{
\textbf{The shared dynamics reconstruction method.} An unobserved driver influences an ensemble of response systems, each containing unique internal dynamics and random measurement filters. Here, we use the R\"ossler dynamical system to drive an ensemble of Lorenz systems with random parameters that have been filtered with random Gaussian response functions. Timepoint-wise weighted recurrence networks are separately calculated for each response using using topological data analysis, and then aggregated to produce a consensus graph. This graph is traversed with either community detection (discrete time) or a diffusive flow (continuous time) in order to reconstruct the driver dynamics.
}
\label{overview}
}
\end{figure*}

Diverse biological and engineered systems exhibit descending control, in which a central driving signal causally influences response variables---which cannot, in turn, influence the driver. In ecology, environmental changes trigger complex cascades across food webs \cite{rosenblatt2016climate}; while in neurobiology, descending neurons convey cognitive imperatives to motor circuits that implement behavior \cite{cande2018optogenetic,vyas2020computation,nande2022bottlenecks}. 
In systems biology, master regulators such as the Hox gene family orchestrate numerous complex developmental processes \cite{leisner2008stochastic}. The ubiquity of top-down control may arise from robustness and parsimony, because it reduces the number of dynamical interactions needed to steer a high-dimensional dynamical system \cite{pezzulo2016top}. However, top-down control introduces experimental challenges when the driving variable is not known {\it a priori}, or is not directly measurable. 

Reconstructing an unknown \textit{driver}, given time-resolved observations of one or more \textit{response} variables, therefore represents a fundamental computational problem with diverse applications in biology and engineering. 
This problem is ostensibly ill-posed, because natural systems dissipate information through noise and other irreversible processes.
However, the theory of dynamical systems provides circumstances in which partial information about an unknown driver may, in principle, be recovered.
Conceptually, a driving signal continuously encodes information about itself into the responding systems. Driver-response coupling therefore acts as a lossy transmission channel, motivating recent causal detection methods grounded in information-theoretic quantities like the transfer entropy \cite{schwarze2022network,runge2012escaping,runge2019detecting}. Alternatively, causal relationships can be inferred based on properties of the dynamical system that generates a time series: theorems from nonlinear dynamics show that partial information present in observable subsystems reveal subregimes within the driving signal \cite{kantz2004nonlinear}, motivating methods that determine causality based on attractor reconstruction or unsupervised learning of latent representations \cite{munch2022recent,sugihara2012detecting,runge2019inferring,de2020latent,bennett2022rethinking,quiroga2000learning,benkHo2018complete}, building upon classical observability conditions for dynamical systems \cite{takens1981detecting,packard1980geometry,stark1999delay}.

However, many existing methods detect only the average strength and directionality of causal interactions among known variables. The more general, and difficult, task of finding time-resolved causal forces remains an open problem. Yet understanding how causal forces change over time is necessary to understand control schemes, particularly in biological systems with sparse or infrequent control signals \cite{sober2018millisecond,durstewitz2023reconstructing}. Time-resolved driving signals may be detected based on signal denoising or assimilation techniques, or by generalizing statistical causality measures with time-dependent kernels \cite{kantz2004nonlinear,dhamala2008estimating,mastakouri2021necessary}. However, these methods fail to leverage existing theoretical works on skew-product dynamical systems and chaotic synchronization \cite{pecora1997fundamentals,hunt1997differentiable,sauer2004reconstruction}, which provide strong inductive biases for physically-motivated representations, even given highly-corrupted measurements. 
As a result, existing temporal causality methods require large datasets, and fail to provide physical insight---when is driver reconstruction possible, and where is this information stored in the response signals? 

Here, we show that recurrences---brief intervals in which time series appear to repeat themselves---represent key dynamical motifs that encode information about unobserved variables driving a time series. Using this finding, we introduce a general-purpose, physics-based unsupervised learning algorithm that reconstructs an unknown driver time series from multiple observed response time series. Based on the intuition that response recurrences imply driver recurrences, we convert time series to networks using tools from computational topology. Traversing this graph reveals temporal progressions of driver states, even from highly nonlinear or noisy response measurements. Our method applies to both discrete- and continuous-time signals and both oscillatory or aperiodic driving, and we show that its underlying accuracy stems from an ergodicity-breaking percolation transition in the time series adjacency graph. We extensively compare our technique to existing methods based on mutual information, artificial neural networks, and spectral transformations, and find that our method obtains strong results on diverse datasets spanning electrophysiology, hydrodynamics, and systems biology. We show that our method's accuracy arises from its ability to identify and approximate unstable periodic orbits that the driving signal transiently shadows---linking our algorithm's performance to topological properties of the underlying dynamical systems.

\section{Approach}

\subsection{Recurrence structure of skew-product causal systems}

Suppose an unknown signal $\v{z}(t)$ drives $N$ response dynamical systems $\dot{\v{y}}_k(t)= \v{g}_k(\v{z}(t), \v{y}_k(t), t)$, $k \in \{1, 2, ..., N\}$. The response signals  $\{\v{x}_k(t)\}_N$ comprise nonlinear measurements $\v{x}_k(t) = \mathcal{M}_k \circ \v{y}_k(t)$. We seek to reconstruct the original driver $\hat{\v{z}}(t)$ given $\v{x}_k(t)$. 

This problem is underdetermined when the measurement operator $\mathcal{M}_k$ is noninvertible, or when each response variable $ \v{y}_k$ only very weakly couples to the driver. However, if the driver state lies on an attractor $\mathcal{Z}$, while the $k^{th}$ response signal lies on an attractor $\mathcal{Y}_k$, then the overall dynamics have {\it skew product}  structure with combined attractor $\mathcal{Z} \times \mathcal{Y}_1 \times \mathcal{Y}_2 \times ... \times \mathcal{Y}_N$. While individual $\mathcal{Y}_k$ are not directly observable, the skew product reconstruction theorem suggests that a one-to-one mapping $\hat{\mathcal{Y}}_k \sim \mathcal{Y}_k$ may be constructed from the observed signal $\v{x}_k(t)$ by applying a lift $\phi(.)$ to the signal using time delays $\hat{\v{y}}_k(t) = \phi(\v{x}_k(t)) = [\v{x}_k(t), \v{x}_k(t - \tau), \v{x}_k(t - 2 \tau), \ldots, \v{x}_k(t - (D-1) \tau)]^\top$, where $D$ and $\tau$ are hyperparameters chosen using a variety of heuristics \cite{takens1981detecting,packard1980geometry,stark1999delay}. Nonlinear generalizations of the lifting map $\phi$ may be obtained with attractor reconstruction techniques like singular value decomposition, kernel methods, or autoencoder neural networks \cite{kantz2004nonlinear,wehmeyer2018time,gilpin2020deep,bakarji2022discovering}.

Recent work by Sauer shows that the one-to-one mapping between $\mathcal{Y}_k$ and $\hat{\mathcal{Y}}_k$ requires that a recurrence in the $k^{th}$ reconstructed response $\hat{\v{y}}_k(t) \approx \hat{\v{y}}_k(t')$, $t \neq t'$ implies a recurrence in the driving signal $\v{z}(t) \approx \v{z}(t')$; however, the converse does not hold true due to the skew product structure \cite{sauer2004reconstruction}. Thus, tracking recurrence events in a time series driven by an ergodic dynamical system gradually reveals information about the driver attractor  $\mathcal{Z}$ \cite{quiroga2000learning}. Longer observations or additional independent response measurements both provide more opportunities for recurrences.

Sauer distills this insight into an exact algorithm for reconstructing a discrete-time periodic driver given noiseless responses. Given a reconstructed response time series $\hat{\v{y}}_k(n)$, $n \in 1, 2, ..., T$, then near-recurrences correspond to two timepoints $n$ and $n'$ where $||\hat{\v{y}}_k(n) - \hat{\v{y}}_k(n') || \leq \epsilon$, where $\epsilon$ is a small constant acting as an adjustable hyperparameter for the method. If a given response series $\hat{\v{y}}_k$ recurs at a set of times $\mathcal{R}_k = \{t^{(k)}_1, t^{(k)}_2, ...\}$, and another response series $\hat{\v{y}}_\ell$ recurs at a set of times $\mathcal{R}_\ell = \{t^{(\ell)}_1, t^{(\ell)}_2, ...\}$, then if any timepoints are shared between $\mathcal{R}_k$ and $\mathcal{R}_\ell$, then {\it all} timepoints in either set belong to the same equivalence class and thus correspond to the same unique driver state. As a result, each one of the $T$ timepoints observed across the $N$ response time series can be labelled by their equivalence class, resulting in a time series of driver states.

However, while this algorithm is motivated by fundamental properties of skew-product systems, it cannot readily be applied to most experimental time series because it requires discrete-time, low-period, and noise-free signals where each observed timepoint can unambiguously be assigned to a single equivalence class. Any spurious recurrences (e.g. from noise) violates the disjointedness of equivalence classes, precluding reconstruction.

\subsection{SHREC: Generalized shared recurrence detection for real-world time series}

We observe that classical skew product reconstruction discovers disjoint sets within an aggregated time series adjacency graph: each lifted response time series $\hat{\v{y}}_k(i) \in \mathbb{R}^{T \times D}$ has distance graph $d_{ij}^{(k)}  \equiv \| \hat{\v{y}}_k(i) - \hat{\v{y}}_k(j) \|$. Classical exact reconstruction merges these $N$ graphs into a consensus graph $d^* \in \mathbb{R}^{T \times T}$, $d^*_{ij} = \inf_k d^{(k)}_{ij}$ and then thresholds at $\epsilon$ to produce the binary adjacency matrix $A_{ij} = \Theta(\epsilon - d^*_{ij})$. Driver state identification therefore requires graph partitioning, a difficult combinatorial optimization problem \cite{mezard2009information}. Strict partitioning precludes detection of continuous drivers, and renders the method fragile to spurious recurrences induced by noise.

Our key insight is that skew product reconstruction describes the long-time behavior of an ensemble of random walkers navigating the time series adjacency graph. Distinct driver states induce kinetic partitions associated with discrete basins of attraction, or equivalence classes. Exact skew product reconstruction therefore represents a low-temperature limit of a more general set of algorithms for analyzing time series. We refer to the resulting class of algorithms as SHREC (\textbf{Sh}ared \textbf{Rec}urrences).

\vspace{2mm}
\noindent\textbf{Improving adjacency estimation.} We first note that identification of recurrence events may be generalized with more robust criteria for recurrence. Rather than setting a fixed distance threshold $\epsilon$ for recurrence, an adaptive threshold would compensate for uneven spacing of timepoints, or nonuniform density of the underlying dynamical attractor \cite{linot2020deep}. Here, we opt to construct a more robust connectivity matrix using a higher-order network embedding method recently introduced for topological data analysis \cite{mcinnes2018umap}. For the $i^{th}$ timepoint of the $k^{th}$ response time series, we first determine the distance to its $M$ nearest neighbors $\{d_{im}^{(k)}\}_{m=1}^M$. We compute the closest neighbor's distance $\rho_i \equiv \min_m d_{im}$, and then numerically solve for the value $\sigma_i$ such that $\log_2 M = \sum_{i=1}^M \exp(-\text{ReLU}(d_{im} - \rho_i) / \sigma_i)$, where $\text{ReLU}(x) \equiv \max(0, x)$. The calculated $\rho_i$, $\sigma_i$ values then transform the distance matrix $d_{ij}^{(k)}$ into an affinity matrix $A_{ij}^{(k)} = \exp(-\text{ReLU}(d_{ij}^{(k)} - \rho_i) / \sigma_i)$. When calculated in this manner, $A_{ij}^{(k)}$ encodes a fuzzy simplicial complex capturing the local density of the attractor $\hat{\v{y}}_k$ \cite{mcinnes2018umap}. Unlike static recurrence plots based on $d_{ij}^{(k)}$, the learned parameters $\rho_i$, $\sigma_i$ modify each row of $d_{ij}^{(k)}$ in a context-dependent manner. Having computed fuzzy simplicial complexes for each individual response, we find the consensus recurrence graph $A_{ij} = (1/K) \sum_k A_{ij}^{(k)}$. 

We use this particular adjacency calculation in our subsequent experiments. We choose to use simplicial complexes due to their empirical success \cite{mcinnes2018umap} and small number of trainable parameters. In experiments, we find that simplicial complexes are more robust to noise than recurrences detected at a fixed threshold $\epsilon$ (\ref{app_simplex_recur}). In general, other heuristics may be used to construct $A_{ij}$, depending on prior information available about the time series, underlying attractor, or dataset-dependent factors. For example, works on simplex-based forecasting scale distances to neighbors $d_{ij}$ using a fixed exponential kernel, implicitly assuming uniform density of the underlying attractor  \cite{sugihara2012detecting}. Other works on driver reconstruction calculate a geodesic distance among points, implying that the original embedding distances are less informative than the graph traversal distances \cite{hirata2008reproduction,nomura2022superposed}. More generally, calculation of $A_{ij}$ represents a metric learning problem, which can be learned directly from data with a probabilistic learning model such as a variational autoencoder, if sufficient data is available \cite{gilpin2024generative}. 
\\

\noindent\textbf{Generalizing equivalence classes for discrete-time drivers.} We next note that identification of dynamical equivalence classes represents a graph clustering problem. A discrete-time driver cycles among a finite set of symbols $s \leq T$, and so driver reconstruction represents the problem of finding a permutation-invariant labeling of $T$ nodes with one of $s$ symbols. We seek a labelling that maximizes connectivity of same-symbol nodes under $A$ while minimizing connectivity between different symbols. For any choice of objective (e.g. modularity, betweenness), optimal assignment of labels represents min-cut, an NP-hard combinatorial optimization problem. Heuristic solutions are provided by community detection algorithms \cite{girvan2002community}, which cluster groups of nodes based on measures of connectivity and similarity within the network.

For our experiments, we use the recently-introduced Leiden algorithm \cite{traag2019louvain}, which balances performance with empirical accuracy in large networks. We find that the performance of our method is insensitive to the choice of community detection algorithm (\ref{app_community}). The Leiden algorithm has time complexity $\sim\!\mathcal{O}(E)$, where $E$ is the number of nonzero edges in $A$. However, for some datasets, custom community detection algorithms could be used instead. For example, for noiseless periodic driving, the recurrence graph approaches a regular structure, for which specialized algorithms detect communities more efficiently \cite{lancichinetti2009community}. Likewise, if certain timepoints are known to correlate strongly with particular driver states (as may occur due to experimental interventions), supervised community detection algorithms may be more accurate.
\\

\noindent\textbf{Continuous-time drivers.} We next note that the concept of dynamical equivalence classes may be generalized to continuous-time driving. Instead of cycling among a finite set of states, a continuous-time driver traverses a continuous manifold of proximal states. To smoothly partition the recurrence graph, we first compute the graph Laplacian $L \equiv A - D$, where $D_{ij} = \sum_k A_{ik}$.  We next compute the subleading eigenvector of $L$, resulting in a continuous labelling of the $T$ nodes in the consensus recurrence graph $A_{ij}$.

In our experiments, we identify driver states using spectral decomposition of the graph Laplacian via Arnoldi iteration. This approach is the most general, because $A$ is not necessarily symmetric (a consequence of using fuzzy simplicial complexes instead of traditional recurrence matrices); but it precludes the use of faster methods like Lanczos iteration. This method also has the same $\sim\!\mathcal{O}(E)$ time complexity as discrete driver reconstruction. However, for specialized datasets, our approach could be generalized by first transforming the graph Laplacian; for example, powers of the Laplacian would highlight slower-timescale driver dynamics. Similarly, pruning the recurrence network may improve driver estimation in systems with intermittent control.
\\

\noindent\textbf{Physical Interpretation.} Both the discrete- and continuous-time driver reconstruction procedures describe the long-time behavior of random walks on the consensus graph $A$. The consensus graph $A$ defines the state transition matrix of a discrete- or continuous-time Markov chain, which propagates distributions of random walkers initialized at different nodes. Repeated application of this operator evolves the walker distribution towards a stationary distribution associated with its largest eigenvalue. If the graph fully partitions into disconnected subgraphs, walkers remain kinetically trapped within their initial basins, revealing an invariant measure associated with the driver. This measure provides the equivalence classes in exact driver reconstruction. However, for most real-world datasets containing noise or corrupted measurements, the transition matrix fails to exhibit fully-disjoint structure due to spurious recurrences that induce leakage among basins associated with distinct driver states. The leading eigenvector of the transition matrix instead approaches the uniform distribution at long times. However, driver structure persists within transients associated with subleading eigenvectors---this information disappears at long times because ergodicity is not fully broken. These weaker partitions are known as "almost-invariant sets" in fluid dynamics \cite{froyland2009almost}, and they correspond to subregions within the graph where random walkers become trapped for extended durations before slowly escaping into other subregions \cite{brunton2017chaos,mezic2005spectral,costa2021maximally}. Recurrence networks thus represent a natural representation of time series, providing a strong inductive bias for downstream analysis tasks.

We demonstrate the key steps of our method in Figure \ref{overview} by using the dynamics of the chaotic R\"ossler attractor to drive multiple realizations of the Lorenz equations with random dynamical parameters. In order to make the reconstruction task more difficult, we filter the Lorenz response dynamics through random Gaussian kernels, resulting in sparse, heavily-corrupted measurements that confound linear reconstruction methods like principal components analysis.

\subsection{Related work and limitations}
Generalizations of time-delay embeddings for non-autonomous systems have previously been used to infer causal structure using likelihood-based methods, particularly for ecological time series \cite{rogers2020hidden,deyle2016global}. A variety of methods use attractor reconstruction to detect causal forces \cite{munch2022recent,sugihara2012detecting,runge2019inferring,de2020latent,bennett2022rethinking,quiroga2000learning,benkHo2018complete}; other works use recurrence events and topological motifs within time series \cite{davidsen2008networks,myers2019persistent}. However, these prior methods determine the average strength and direction of causal interactions between time series, while our method instead seeks a time-resolved driver. Among time-resolved methods, our algorithm conceptually resembles prior methods that aggregate recurrence graphs, and then identify contiguous paths through the resulting network using the Isomap algorithm \cite{hirata2010identifying,donner2010recurrence,hirata2008reproduction,nomura2022superposed,tenenbaum2000global}; however, instead of static recurrence graphs we use a consensus graph of fuzzy simplicial complexes, and unlike prior works our traversal method has a dynamical interpretation as a diffusive flow (for continuous-time drivers) or jump process (discrete-time). Our reconstruction process conceptually resembles pseudotime, a topological graph ordering method used in systems biology to infer developmental trajectories from gene expression measurements of unsorted cell populations \cite{haghverdi2016diffusion}. Other works note that generalized synchronization between two time series reveals fixed causal graphs \cite{rulkov1995generalized,pecora1997fundamentals,hunt1997differentiable,huang2020detecting}. 
When the exact model class of the driver is known, generalized synchronization further reveals the full temporal driver dynamics, because temporal driver reconstruction requires only identifying a finite number of variables (like phase and frequency) that parameterize the driver \cite{yu2008dynamical,chen2007chaos,quinn2009parameter}. Instead of treating each response variable as a dynamical system, recent works on temporal Bayes filtration and neural data assimilation treat response observations as instantaneous nonlinear samples from a latent process \cite{mccabe2021learning,frerix2021variational,karl2016deep,krishnan2015deep,koppe2019identifying,hamilton2016ensemble}. Our emphasis on incremental information gain from recurrences distinguishes our approach from multiview and cross-embeddings \cite{ye2016information,leng2020partial,munch2022recent,runge2019inferring,de2020latent,bennett2022rethinking,quiroga2000learning,benkHo2018complete}, as well as from generalizations of static causal network inference to temporal networks \cite{alaa2019validating,mastakouri2021necessary,williams2022shape,kantz2004nonlinear,dhamala2008estimating}. 

In contrast to strong causality associated with interventional studies and counterfactual analysis \cite{pearl2010causal}, our work detects weak causality as skew product coupling in settings where only observational data is available. Like other observational causal discovery methods \cite{pecora1990synchronization,rogers2020hidden,deyle2016global,hirata2010identifying,runge2019inferring}, our approach bears restrictions regarding the \textit{identifiability} of the driving signal: in the absence of interventions, it is impossible to differentiate among sets of potential causal drivers that exhibit equivalent influences on the dynamics. For example, for the system $\dot x = x + h(z(t - \tau))$, the true driving signal for $x(t)$ could be defined as $z(t)$, $h(t)$, $(t - \tau)$ or $h(t - \tau)$. Likewise, given a multivariate driving signal $\v{z}(t) = z_1(t), z_2(t), ...$, it is impossible to distinguish among rotations and combinations of potentially independent causal coordinates, though overembedding with time-delays mitigates some ambiguity \cite{hegger2000coping}. While our method readily accommodates driver-synchronized response systems $x_i(t) = h_i(z_i(t))$, it requires mutual desynchronization among a subset of response systems, $x_i(t) \neq h_i(x_j(t))$ for some $i\neq j$, in order to avoid the degenerate solution $\hat{z}(t) = x_i(t)$. 

\section{Results}

\subsection{The shared recurrences algorithm identifies causal drivers in diverse datasets}

\begin{figure*}
{
\centering
\includegraphics[width=\linewidth]{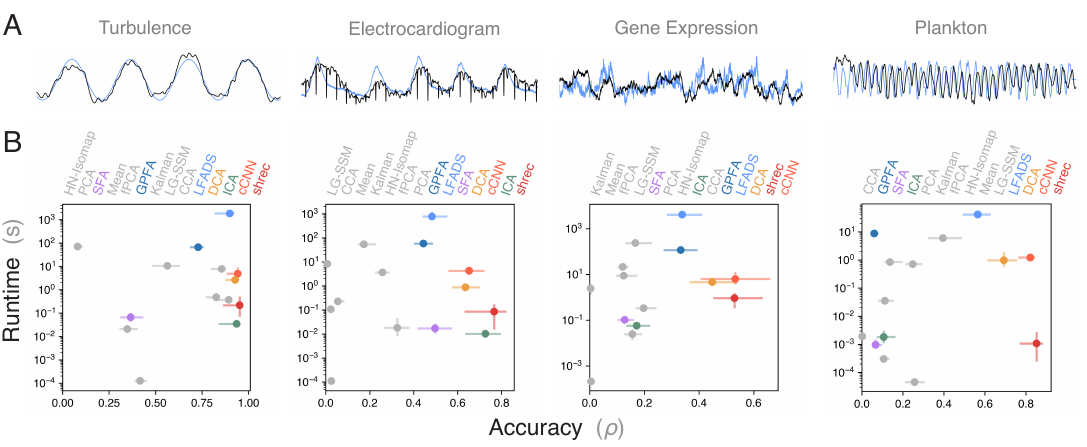}
\caption{
{\bf Applying the shared recurrences (SHREC) algorithm to diverse datasets.} (A) Example driver signals (blue) and their reconstructions (black) from response variables for four different experimental time series. (B) Runtimes versus Spearman correlation scores for different reconstruction methods. Error bars denote bootstrapped one-sided standard errors around random model initializations trained on random subsets of the original time series. The different methods sorted by overall score are annotated above each panel.
}
\label{explore}
}
\end{figure*}

We first perform large-scale benchmarks that demonstrate the practical advantage of our approach. We will show that our physics-based approach, which reduces to exact dynamical equivalence classes in the limit of zero noise, exceeds the performance of simpler methods, while matching the performance of less-interpretable methods based on artificial neural networks. Rather than highlighting raw performance, we aim to emphasize how a physically-motivated, recurrence-based approach provides a strong inductive bias for time series decomposition, allowing strong results to be obtained from a relatively simple computation.

Our benchmark experiments consist of training our method and other methods on only the response time series ${\v{x}_k(t)}$, in order to produce an estimate of the driver $\hat{\v{z}}(t)$ for each benchmarked method. We compare each reconstruction to the ground truth driver $\v{z}(t)$ to assess relative method performance. We emphasize that our technique is an \textit{unsupervised} learning method: in a real-world setting, no ground truth $\v{z}(t)$ is available, underscoring the importance of benchmarks and validation on datasets with known drivers.

For our benchmarks, we consider a diverse range of datasets with causal structure.\\
\textbf{Turbulence:} Trajectories of diffusing tracer particles advected by a chaotic double gyre flow while also experiencing Brownian motion. The driver is long-wavelength sinusoidal forcing, and the responses are the radial coordinates of individual particles (as would typically be observed in Doppler velocimetry). \\
\textbf{ECG:} A fetal electrocardiogram. The driver is the maternal respiration signal, while the responses are multi-point electrocardiogram recordings placed across the fetal body \cite{sulas2021non}.\\
\textbf{Gene Expression:} Stochastic gene expression dynamics in a subset of the yeast metabolic gene regulatory network, as generated by a microbiology experiment design suite \cite{schaffter2011genenetweaver}. The driver is the expression level of a known transcriptional regulator in after a knockdown perturbation, and the responses are the expression levels of downstream genes. These time series represent longer versions of the those used in DREAM4, a standard benchmark for regulatory network inference \cite{marbach2010revealing}.
\\
\textbf{Plankton:} Plankton species abundances in an alpine lake in Switzerland over a $20$ year period, as the ecosystem undergoes temperature fluctuations \cite{pomati2020interacting}. The driver is the monthly average temperature, and the response variables are individual plankton species abundances. $30\%$ of all values in this dataset are missing.

We compare our method to a variety of alternative models: principal components analysis (PCA), independent components analysis (ICA), canonical correlation analysis between past and future values (CCA), simple averaging (Mean), a Kalman filter (Kalman), and a surrogate-weighted ensemble Fourier transform (fPCA)\cite{kantz2004nonlinear}. Recently-introduced dynamical decomposition methods include dynamical components analysis (DCA), which finds low-dimensional modes that maximize predictive information \cite{clark2019unsupervised}; slow-feature analysis (SFA), which seeks linear combinations of nonlinear kernels that minimize time variation \cite{wiskott2002slow}; Gaussian-process factor analysis (GPFA), which finds latent factors parametrizing the mean of a one-step correlated Gaussian process \cite{yu2008gaussian}; and a linear Gaussian state space model (LG-SSM) \cite{sarkka2013bayesian}. Another recent causality detection approach (HN-Isomap) first thresholds and merges each response's traditional recurrence graph (not the context-dependent recurrence matrix used here) and then computes a weighted affinity graph using Dijkstra’s algorithm \cite{hirata2008reproduction,nomura2022superposed}. We also consider artificial neural networks in the form of a causal convolutional network (cCNN), an established architecture for time series representation learning that trains a custom convolutional filter on the response time series \cite{franceschi2019unsupervised}; and Latent Factor Analysis via Dynamical Systems (LFADS), a recurrent variational autoencoder \cite{pandarinath2018inferring}. We do not directly benchmark our method against Sauer's original exact algorithm, which only applies to noise-free discrete-time datasets; on the real-world datasets considered here, Sauer's method produces a monotonically-increasing trendline, because nearly every timepoint is assigned to a new equivalence class. 

For all benchmark models we tune hyperparameters using cross-validation and grid search on a separate training dataset than the held-out testing data. We do not vary any hyperparameters for our proposed method, SHREC. We use $8$ accuracy metrics to compare the reconstructed driver $\hat{z}(t)$ to the true driver $z(t)$: mean-squared error, Spearman correlation, covariance, symmetric mean absolute error, mean absolute scaled error, dynamic time warping distance, mutual information, and nonlinear Granger causality; all yield comparable results, and so we report Spearman correlations for simplicity. Additional metrics and benchmark experiments are included in \ref{supp_benchmarks}.

We observe strong performance of our approach relative to other methods---both conceptually and computationally simpler methods such as PCA or SFA, but also artificial neural networks like LFADS (Fig. \ref{explore}). A comparably strongly-performing benchmark model, the causal autoencoder neural network, takes the hierarchical and sequential structure of time series data into account \cite{franceschi2019unsupervised}, but otherwise does not specifically exploit skew product structure in the underlying time series. Additionally, because our method requires only matrix decomposition and not hyperparameter tuning or iterative gradient descent, it ranks among the fastest methods to compute relative to its performance. Our method performs particularly well on the $30\%$-corrupted plankton dataset. Our method consistently appears among the highest-performing methods across all datasets, a particular strong result given that our approach is purely physics-based---it reduces to exact dynamical equivalence classes in the limit of noise-free, discrete-time driving. Our benchmark results therefore highlight the strong inductive bias of our method for skew-product systems.

Across the experimental datasets, we find that aperiodic driving signals are consistently harder to reconstruct by all methods, with the gene expression and plankton datasets showing the lowest score ranges. The presence of multiple significant but widely-spaced timescales presents a more challenging problem setting, which motivates our exploration of unstable periodic orbits in Section \ref{sec:upo}. Interestingly, while the turbulence dataset contains a monochromatic periodic driver, linear methods like the Fourier transform or PCA are unable to detect this signal due to relatively high noise in the observed data. Many of our example datasets represent cases in which the reconstructed driving signal $\hat{z}(t)$ exerts nearly-instantaneous influence on the response dynamics, $\hat{z}(t) \sim z(t)$. Known as the synchronized regime in prior work \cite{monster2017causal,ye2015distinguishing}, this regime is most appropriate for benchmarks both because of the causal identifiability limitation (discussed above), and because the synchronized limit represents the most competitive setting for instantaneous latent variable models like PCA and SFA.

\subsection{The shared recurrences method identifies higher-order interactions among time series}

\begin{figure}
{
\centering
\includegraphics[width=\linewidth]{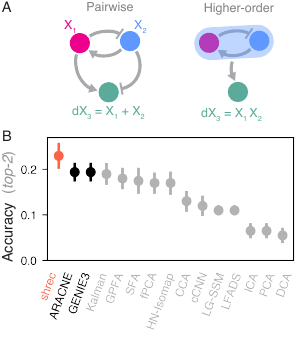}
\caption{
{\bf Shared reconstruction identifies higher-order interactions.} (A) Dynamics driven by pairwise interactions versus a single multiplicative driver. (B) Accuracy of network inference methods in identifying two multiplicative driving nodes among a network of $N$ interacting genes. Error ranges are standard errors across $100$ random networks with one multiplicative driver pair among $N=12$ genes.
}
\label{simplex}
}
\end{figure}

We next consider a setting in which SHREC recovers drivers inaccessible to traditional time series analysis methods.
Many biochemical circuits contain higher-order interactions, in which combinations of upstream factors jointly regulate downstream targets \cite{bick2023higher,sanford2020gene}. Unlike traditional approaches to observational causality detection, in which a graph fully captures the relationship among time series, higher-order terms require a hypergraph encoding relationships among combinations of nodes. This problem has seen little prior attention in the context of causal reconstruction, but it bears particular relevance to gene expression, in which individual genes' temporal (or pseudotemporal) dynamics are directly observable, but the regulatory effects of combinations are not. We thus expand our gene expression example from the previous section, by including higher-order terms, in which the product of two genes drives each downstream gene's dynamics. This multiplicative interaction appears in multi-input motifs, a common feedback mechanism in transcription networks \cite{sanford2020gene}. However, few approaches to gene regulatory network inference directly detect it. Here, we show that the purely physics-based driver reconstruction of SHREC identifies higher-order information, allowing it to outperform both baseline causality methods and specialized network inference methods in identifying higher-order drivers.

We modify an established approach to simulating dynamics of gene networks, in order to include higher-order interactions (see \ref{simplex_app}). We simulate a set of $N$ genes, in which the product of two genes regulates the dynamics of the remaining genes by random positive or negative amounts. We then apply SHREC to the set of $N$ time series, and we then correlate the resulting reconstructed driver $\hat{y}$ with each of the $N$ individual gene time series, and denote the top $2$ genes as the identified regulator genes. We report the accuracy as the fraction of the top $2$ most correlated genes that correctly match the true upstream genes. However, while we use this accuracy metric to benchmark our approach, the results do not change if we instead assess quality by directly comparing the reconstructed driver and true multiplicative driver using the same metrics used in the previous section's benchmarks (\ref{simplex_app}). However, using a node-based accuracy criterion allows us to further benchmark our approach against two standard baseline gene regulatory network inference techniques, GENIE3 and ARACNE \cite{huynh2010inferring,margolin2006aracne}. These methods infer an $N \times N$ interaction matrix given a set of $N$ time series. For these algorithms, we use the two nodes with the highest outdegree as the method's estimate of the driver.

We find that SHREC matches or exceeds the performance of other methods in identifying the genes responsible for higher-order driving (Fig. \ref{simplex}). The simple, physics-based approach of SHREC thus allows it to perform surprisingly well on established problem in network science with particular relevance to bioinformatics. 
Many standard network inference methods, including GENIE3 and ARACNE, detect statistical dependencies among genes in a network, and ignore the dynamical context available in time series \cite{kernfeld2024transcriptome}. 
In contrast, SHREC has a strong inductive bias for dynamical systems, allowing it to perform well on this problem, where the sequential structure of a time series is informative.
Thus while other, specialized gene network inference methods could potentially perform better, we emphasize that SHREC's strong performance comes without any fine-tuning or specialization.

\subsection{Driver reconstruction precedes effective synchronization}
\begin{figure}
{
\centering
\includegraphics[width=0.7\linewidth]{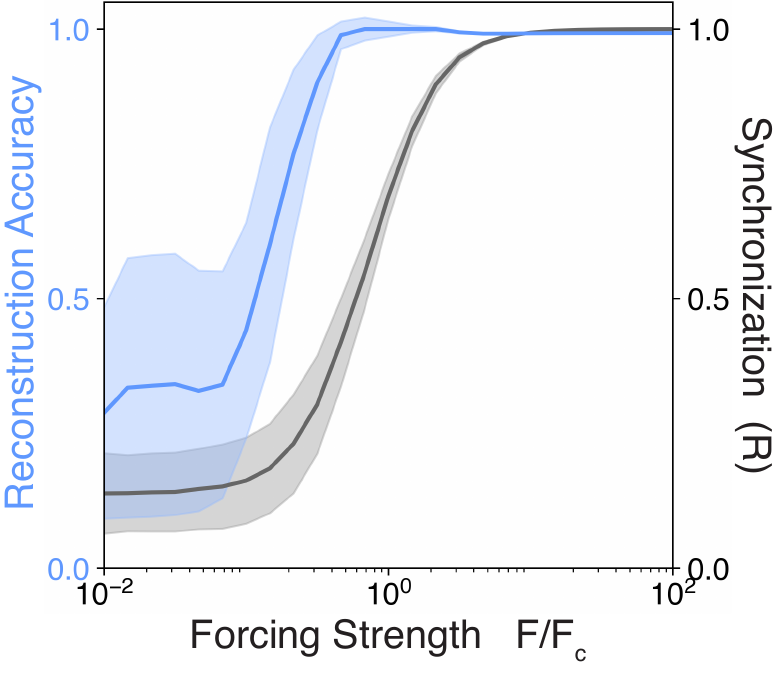}
\caption{
{\bf Driver reconstruction across a synchronization transition.} The Kuramoto order parameter $\langle R(t) \rangle_t$ (black) and the accuracy of driver reconstruction (blue) as the driver forcing strength increases. In this figure, $N=10^3$, $K = \Delta = 1$ in \eqref{kuramoto}, and error bars are standard deviations over $50$ randomly-initialized replicates.
}
\label{sync}
}
\end{figure}

To probe the physical interpretation of driver reconstruction, we consider a synthetic task in which the driver and responses exhibit varying degrees of synchronization. We consider the forced Kuramoto model, a classical model of synchronization among phase oscillators,
\begin{equation}
	\dot\theta_i(t) = \omega_i + \dfrac{K}{N}\sum_{j=1}^N \sin(\theta_j - \theta_i) + F \sin(\omega_f t - \theta_i)
\label{kuramoto}
\end{equation}
where $\theta_i(t)$ is the phase of the $i^{th}$ of $N$ oscillators,  and $F$ is the amplitude of an external driving force with intrinsic frequency $\omega_F$. The frequencies are randomly-drawn from a normal distribution $\omega_i \sim \mathcal{N}(\omega_0, \sqrt{\pi / 2}\Delta)$; when all oscillators are identical ($\Delta=0$), the system always approaches a globally-stable synchronized state $\theta_i(t) = \langle \theta_j(t) \rangle_j$ for all $i$.

When the oscillators are heterogenous ($\Delta>0$), \eqref{kuramoto} exhibits a well-studied synchronization transition as the coupling $K$ increases. When $F=0$, synchronization occurs when $K = 2\Delta$. Analytical reductions of \eqref{kuramoto} for $F \neq 0$ show that, when $K < 2 \Delta$, synchronization instead occurs at a critical coupling strength $F_c$, which is given by a complex but exact expression in terms of $K$, $\Delta$, and the detuning factor $(\omega_f - \omega_0)$ \cite{ott2008low,childs2008stability}. In Fig \ref{sync}, we show this transition at fixed $K < 2\Delta$. We quantify synchronization using the time-averaged Kuramoto order parameter $\langle R(t) \rangle_t$,
\[
	R(t) e^{i \Theta(t)} \equiv \sum_{j=1}^N e^{i \theta_j(t)},
\]
where the synchronized case $\langle R(t) \rangle_t=1$ implies long-term phase-locking ($\theta_i(t) = \langle \theta_i(t) \rangle_i$ for all $i$), while the desynchronizd state $\langle R(t) \rangle_t = 0$ implies that the phases are uncorrelated on average. 

For each value of $F$, we apply the continuous-time version of SHREC to the time series $\{ \theta_i(t) \}_N$, and compute the Spearman correlation between the reconstruction $\hat{z}(t)$ and true driver $z(t) \equiv \sin(\omega_f t)$ (Fig. \ref{sync}, blue trace). As expected, our method readily detects the driver in the synchronized regime, where all responses mirror the driver $\theta_i(t) \propto z(t)$. However, our method also successfully detects the driving signal one decade below the synchronization threshold. This regime corresponds to partial synchronization in the Kuramoto model, in which groups of oscillators temporarily entrain and intermittently approach the driver frequency $\dot\theta_i(t) \rightarrow \omega_f$ without becoming phase-locked. The intermittent coincidence of recurrences due to these temporary entrainments provides a noisy signal of the driver, providing sufficient information for the shared dynamics method to recover it.

\subsection{Chaotic drivers hinder reconstruction, but chaotic responses improve it}
\begin{figure}
{
\centering
\includegraphics[width=\linewidth]{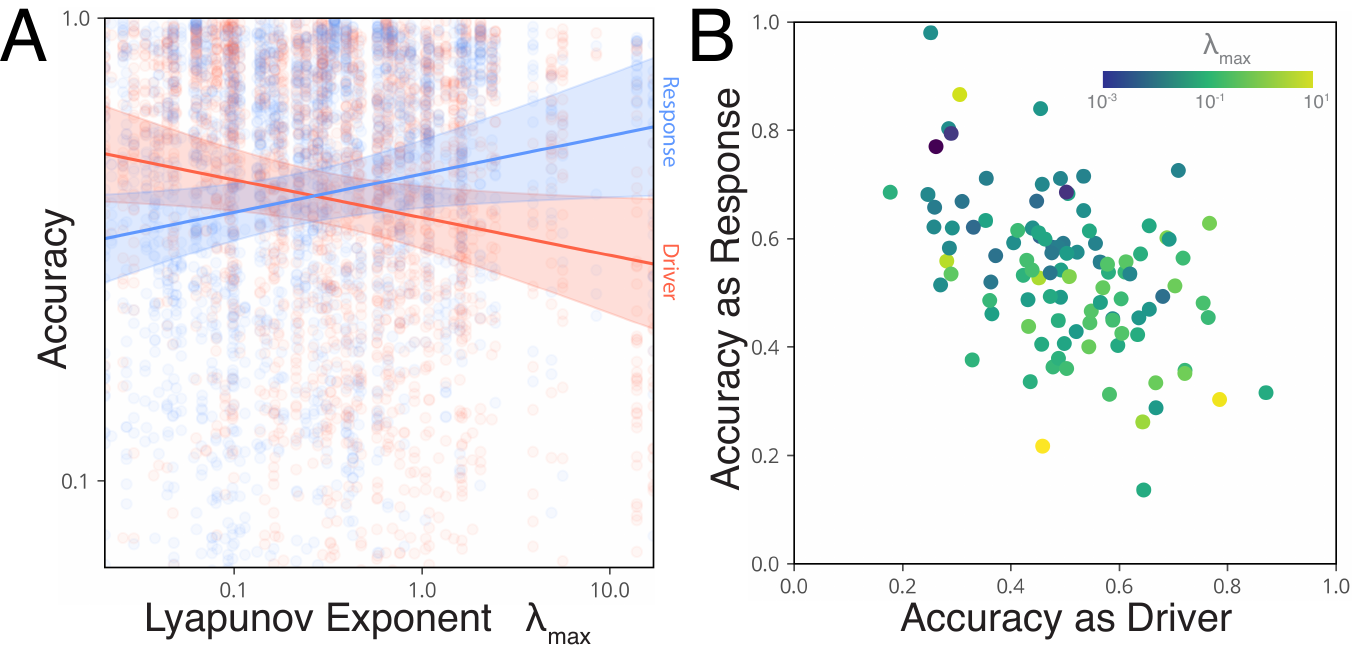}
\caption{
{\bf Correlation between chaoticity and accuracy across thousands of dynamical systems.} (A) Accuracy of driver reconstruction across thousands of random skew-product dynamical systems. Each system is produced by sampling pairs from a set of $135$ named chaotic systems (e.g. Lorenz, R\"ossler, etc) and using one system as a driver for replicates of the other system. Linear fits with 95\% bootstrapped confidence intervals highlight significant Spearman correlations between the reconstruction accuracy and the Lyapunov exponents of the driver (red, $\rho=0.16 \pm 0.03$) and response (blue,  $\rho=-0.20 \pm 0.03$). (B) For each individual dynamical system, the average accuracy of a reconstruction in which it appears as a response variable, versus one in which it appears as a driver.
}
\label{chaos}
}
\end{figure}

In order to further understand which systems are more difficult to reconstruct, we perform an additional benchmark using our recently-introduced database of $135$ low-dimensional ordinary differential equations \cite{gilpin2021chaos}. Curated from published literature to include common systems like the Lorenz, R\"ossler, and Chua attractors, the dataset has grown through crowdsourcing to span diverse domains such as astrophysics, climatology, and biochemistry. Each system is annotated with calculations of descriptive mathematical properties such as maximum Lyapunov exponents, fractal dimensions, and entropy rates, which quantify different aspects of each system's underlying complexity.

We use our dataset to create random skew-product systems: one randomly-selected system is chosen to drive multiple randomly-initialized versions of another randomly-selected system. The Lorenz-R\"ossler system featured in Figure \ref{overview} represents one such skew-product system from the $\sim\!\! \!10^5$ sampled pairs. We scale the coupling strength to a constant multiple of the average amplitude of both systems, and we scale integration timesteps and trajectory lengths based on precomputed estimates of Lipschitz constants and dominant timescales annotated with each system within our database. Of all the possible pairs of systems, we retain only time series from the $5 \times 10^3$ skew product systems that exhibit complex dynamics. For each skew product system, we apply our reconstruction method in order to estimate reconstruction accuracy for that particular driver-response pair. 

Across all pairs, we observe weak but consistent negative correlation between the driver Lyapunov exponent and reconstruction accuracy, implying that more chaotic systems are more difficult to reconstruct (Fig. \ref{chaos}A). Surprisingly, we observe the opposite for response systems: more chaotic response systems improve reconstruction quality. As a result, a particular dynamical system that proves easier to reconstruct, will represent a worse observable for other systems (Fig. \ref{chaos}B).

This discrepancy arises from the dual roles of the driver and response systems: more chaotic response dynamics produce more diverse and thus informative time series, leading to more unique recurrence events. Conversely, a non-chaotic response ($\lambda_\text{max} \approx 0$) yields only a finite amount of driver information over an extended duration, or across many replicates. Less chaotic systems thus hinder reconstruction because they discover new information about the driver at a slower rate. However, more chaotic drivers transmit information at a faster rate, thus requiring more information to be gained from the response time series. This concept of the driver as an information source, and response systems as receivers, mirrors classical formulations of skew-product systems \cite{adler1983algorithms,baptista2005chaotic}. The entropy production rate of a chaotic system is bounded from below by the largest Lyapunov exponent \cite{pesin1977characteristic}, which acts as a proxy for the information transmission rate between the driver and response. In the limit of perfectly-lossless transmission from driver to response system, the systems would exhibit generalized synchronization, in which the response becomes a function of solely the driver state \cite{pecora1990synchronization,verzelli2021learn}. 

\subsection{Accurate reconstruction requires a percolation transition in the recurrence network.}
\label{sec_percolation}

\begin{figure}
{
\centering
\includegraphics[width=\linewidth]{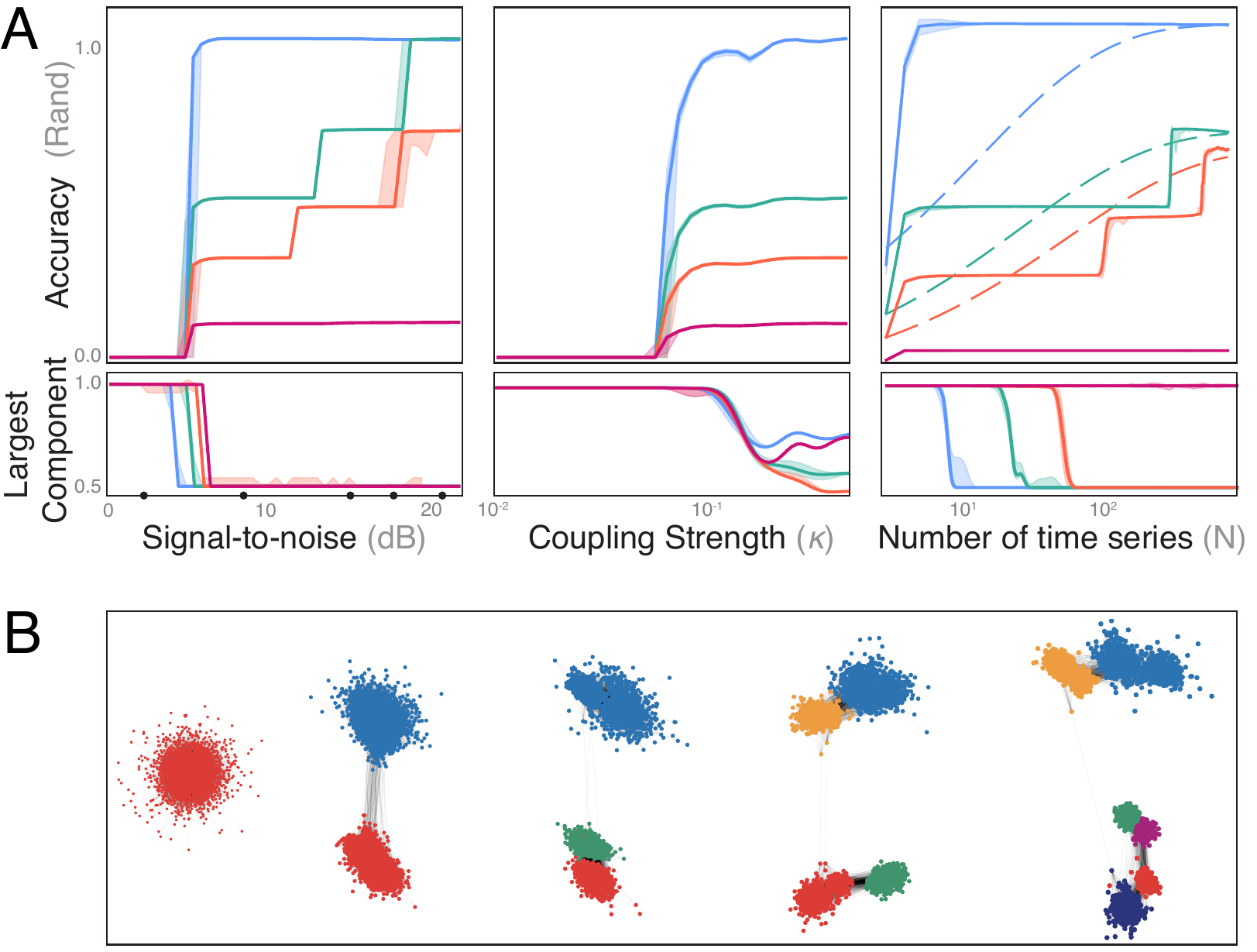}
\caption{
{\bf A period-doubling bifurcation reveals that percolation loss precedes accurate reconstruction.} (A) Driver reconstruction accuracy (adjusted Rand index) for responses driven by the logistic map in its period-2 (blue), period-4 (turquoise), period-8 (red), and fully-chaotic (magenta) regimes; plotted from left to right as a function of the noise level, driver to response coupling strength, and number of response time series. In the last plot, dashed guides indicate null hypotheses for scaling of accuracy with dataset size, as predicted by the $\beta$-distribution. (Lower) The size of the largest connected component (a measure of percolation) in the reconstructed time series adjacency graph. All plots show median and interquartile range across $60$ replicates with random initial conditions and algorithm seeds. (B) Changes in inferred adjacency graph of the period-8 sequence as the data noise is varied; frames correspond to points on the red trace marked with dots in the first panel. 
}
\label{properties}
}
\end{figure}

Our reconstruction method implicitly defines a diffusion process on the recurrence network $A$. In order to understand how this dynamical process couples to the dynamics of the driver itself, we apply the discrete-time variant of our algorithm to systems driven by the logistic map as it undergoes a series of period-doubling bifurcations leading to chaos (Fig. \ref{properties}A). We track the point-wise accuracy of the reconstructed discrete driver using the adjusted Rand index, which accounts for the permutation invariance of labels: an adjusted Rand index of $0$ implies a driver state labelling with no greater accuracy than random chance, while a score of one implies that every driver state is correctly identified for each of the $3 \times 10^3$ timepoints tested. We record our approach's accuracy as a function of three control parameters: the relative amount of stochasticity present in the response dynamics (signal-to-noise ratio), the strength of coupling between the response dynamics and driver, and the number of response time series used for reconstruction $N$. In the latter case, we naively expect that reconstruction accuracy will increase proportionally to the total number of recurrences observed and thus $N$, resulting in an accuracy scaling given by the $\beta$-distribution, which describes the distribution of the expected value of the mean of a binomial process in which driver recurrence states are discovered at random.

We observe that shorter-period drivers are generally easier to reconstruct (Fig. \ref{properties}A). For higher-period drivers, we observe a staircase pattern in the accuracy as reconstruction quality improves. This occurs because, at lower accuracies and longer periods, our approach fails to differentiate all driver states, resulting in groups of states merging together into a coarse-grained, lower-period driving signal. These clusters split into distinct driver substates as reconstruction quality improves (Fig. \ref{properties}B).

In order to understand how changes in method accuracy depend on structural properties of the underlying recurrence graph, for each condition we measure the scaled largest connected component $T_{LCC} / T \in [0, 1]$, where $T$ is the total number of timepoints and $T_{LCC}$ is the number of nodes comprising the largest connected component of $A$. This order parameter detects percolation in finite-size undirected networks \cite{hebert2019smeared}, and in our system it reveals a first-order phase transition preceding an increase in accuracy when any of the three control parameters is varied. Intuitively, a highly-connected time series network contains many equivalent paths between two nodes, producing ambiguity regarding the correct assignment of recurrence events to driver states. This degeneracy below the percolation transition produces an underdetermined reconstruction, which becomes determined as the amount or quality of response datasets increases. Considering a diffusion process on the recurrence graph, increasing any of the three control parameters increases the duration of transients that reveal almost-invariant structure. The percolation transition therefore indicates the onset of weak ergodicity breaking, in which the dynamics transition from having a short mixing time to a much longer time dominated by gradual leakage between highly-connected groups of nodes sharing the same driver state. 

Our observations mirror recent results reporting equivalency between the fitting accuracy of artificial neural networks and the jamming transition \cite{geiger2019jamming}. By decreasing the ratio of data to model size, a neural network passes from underfitting to overfitting regimes mirroring a dilute-to-jammed transition. In our system, increasing training data size or effective size (by decreasing noise or increasing coupling) increases final accuracy by making the time series adjacency graph more dilute, and thus ensuring that distinctions emerge among different driver states. High-period driver states initially appear as low-period states at the onset of the percolation transition, an observation reminiscent of renormalization in period-doubling systems \cite{feigenbaum1979universal}.

\subsection{Reconstruction shadows dominant unstable periodic orbits.}
\label{sec:upo}

\begin{figure}
{
\centering
\includegraphics[width=\linewidth]{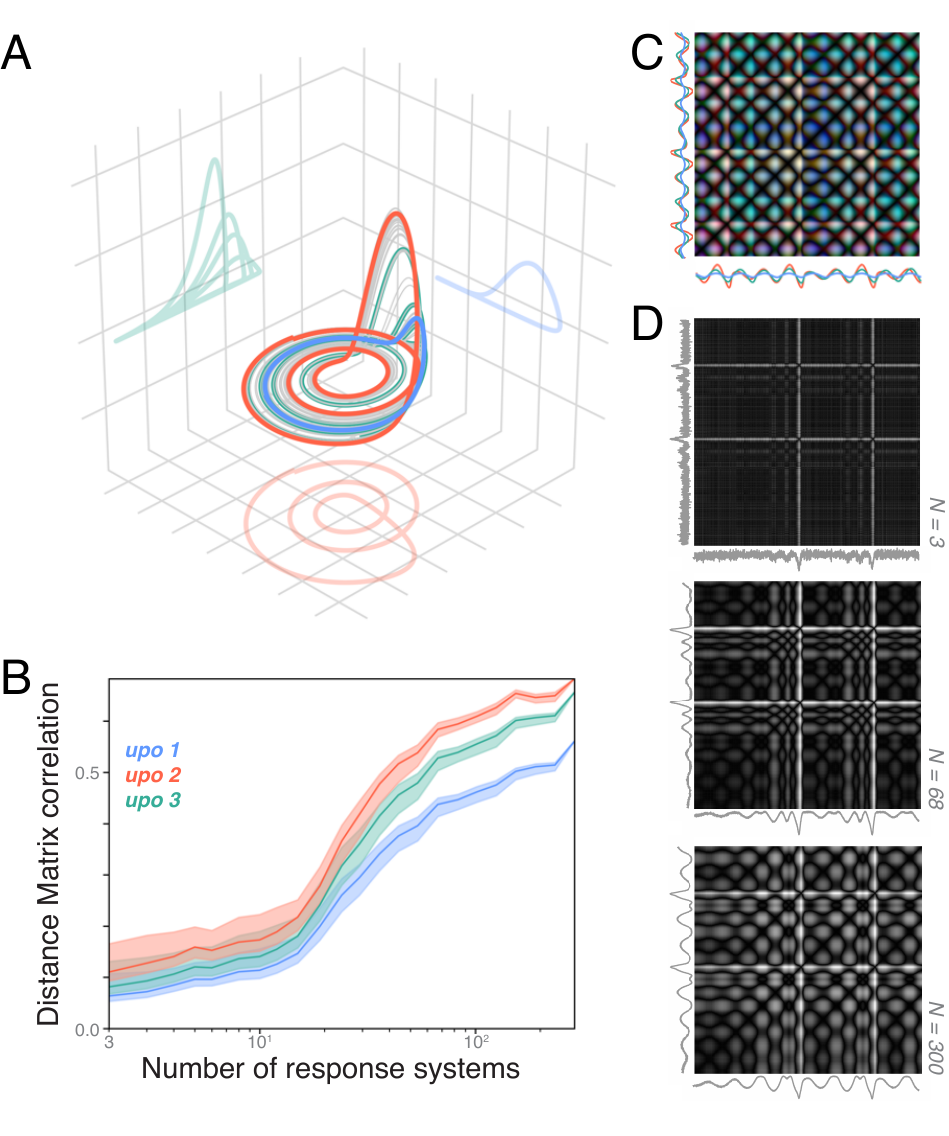}
\caption{
{\bf Reconstruction of complex signals follows dominant unstable periodic orbits.} (A) The three most stable unstable periodic orbits for the R\"ossler system. (B) Elementwise Pearson correlation between the distance matrix of the reconstructed R\"ossler driver and the three orbits, as the number of response time series (and thus overall driver reconstruction accuracy) increases. Ranges denote 95\% confidence intervals around random model initializations trained on random subsets of available response series. (C) The overlaid true distance matrices of the three leading orbits, with RGB channels of the color image encoding each orbit's individual distance matrix. (D) The distance matrices of the reconstruction as the number of response subsystems increases.
}
\label{orbit}
}
\end{figure}

We next consider the process by which our method learns to reconstruct diverse signals. We return to our original demonstration, in which the R\"ossler dynamical equations drive many randomly-corrupted measurements of the Lorenz system. We repeat the reconstruction while varying the number of response series $N$, and thus the reconstruction accuracy (Fig. \ref{orbit}).

Given our method's emphasis on inferring driver recurrence states, we speculate that its accuracy implicitly depends on the unstable periodic orbit spectrum of the driver \cite{hunt1997differentiable}. Classical work on nonlinear dynamical systems has shown that any chaotic system can be decomposed into an infinite set of unstable periodic orbits \cite{cvitanovic1988invariant}, which trajectories shadow for extended durations proportional to each orbit’s relative stability. While chaotic systems have continuous power spectra, the set of valid unstable periodic orbits bears basic topological restrictions due to attractor geometry---making it a countable, albeit infinite, hierarchical measure for the dynamical system. We compute several unstable periodic orbits of the R\"ossler driving system using the method of closest recurrences \cite{lathrop1989characterization}, and select the three most dominant orbits based on the relative duration they are shadowed by the driver dynamics.

Comparing the pairwise distance matrix of the reconstructed signal with the distance matrix of individual unstable periodic orbits suggests that, as the amount of training data increases, the reconstructed distance matrix initially approximates only the most dominant orbit, but then gradually approximates finer-scale details of the driver attractor that are encoded in secondary orbits. We hypothesize that the recurrence structure of our method biases it towards shorter-period and more stable orbits that dominate the graph partition. This phenomenon provides some context for our earlier observation that reconstruction accuracy depends inversely on the driver's Lyapunov exponent: systems with larger Lyapunov exponents transition among a larger set of orbits, thus requiring more recurrence information to produce faithful reconstructions \cite{ott2002chaos}.

\section{Discussion}

We have presented a theoretically-motivated unsupervised learning method that reconstructs an unobserved causal driving signal, given observations of downstream response variables. Across extensive benchmarks, we find that our method quickly and robustly reconstructs diverse drivers from datasets ranging from fluid dynamics to systems biology. Beyond introducing a new data analysis method, our work provides empirical insight into the nature of information transmission in coupled dissipative systems: the topology of the driver's characteristic unstable orbits manifests as structure within the recurrence graph of the response time series, with the efficiency of this linkage mediated by both the chaoticity of the driver, and the diversity of the responses. A diffusion process on the resulting time series graph reveals driver structure in its longest-lived transient modes, and methods that increase the lifetime of these transients---such as more response datasets, increased driver-response coupling, or reduced measurement noise---produce a corresponding increase in reconstruction accuracy.  Diffusion represents a heuristic solution to the difficult combinatorial optimization problem of graph traversal and node sorting, and our method reduces to a previously-known exact algorithm in the limit of a noiseless and discrete periodic driver \cite{sauer2004reconstruction}.

Limitations of our approach therefore stem from our use of diffusion for graph traversal. Certain measurement types may induce false recurrences with systematic structure, such as sensor saturation or memory effects, that impose fixed structural barriers to diffusion on the graph adjacency matrix---akin to inhomogenous diffusivity. Mitigating such effects requires sufficient information about the measurement dynamics to reconstruct and correct for response bias; within our framework, this might require preconditioning the graph Laplacian \cite{krishnan2013efficient}. Additionally, non-stationary driving signals can complicate reconstruction; for example if the attractor is time-dependent (e.g. energy gradually dissipates due to drag), then the driver may exhibit transient chaos before settling into a state of quiescence or oscillation \cite{motter2013doubly}. In this case, a random walker's transitions among nodes will no longer heed detailed balance and the transition operator will exhibit non-Hermitian structure, resulting in long-lived nonequilibrium steady-states that complicate driver state assignment \cite{fruchart2021non,battle2016broken,weis2022coalescence}. These and other, more challenging data types may require nonlinear alternatives to diffusion processes on the recurrence graph. Our use of diffusion therefore represents an inductive bias in the time series representations learned by our model.

However, the problem of discovering nonlinear processes from an empirical recurrence graph introduces a variety of potential generalizations grounded in recent results from statistical learning. Instead of directly fitting a linear operator, future work could leverage recent methods for data-driven inference of dynamical propagators \cite{bramburger2021data,raissi2018hidden,chen2018neural,choudhary2020physics,gilpin2024generative,chen2021data,hamzi2021learning}, which can even produce reduced-order analytical models \cite{champion2019data}. Generalizations may also incorporate alternative graph representation or traversal methods (such as message passing)\cite{williams2022shape,gilmer2017neural,thongprayoon2023embedding}, such as graph neural networks \cite{donoho2009message,rossi2020temporal}.  Thus, beyond a specific algorithm, our work motivates general parametrization of time series in terms of recurrences, and the latent orbit structure that they encode.

\section{Acknowledgments}

We thank Ski Krieger, Denitsa Milanova, Phil Morrison, Anthony Bao, and Rob Philips for helpful discussions. WG was supported by NSF DMS 2436233 and Marble Therapeutics. This project has been made possible in part by grant number DAF2023-329596 from the Chan Zuckerberg Initiative DAF, an advised fund of Silicon Valley Community Foundation. Computational resources for this study were provided by the Texas Advanced Computing Center (TACC) at The University of Texas at Austin.

\clearpage
\renewcommand{\thetable}{S\arabic{table}}
\setcounter{table}{0}
\renewcommand{\thefigure}{S\arabic{figure}} 
\setcounter{figure}{0}
\renewcommand{\theequation}{A\arabic{equation}}
\setcounter{equation}{0}
\renewcommand{\thesubsection}{\Alph{subsection}}
\setcounter{subsection}{0}
\renewcommand{\thesection}{Appendix \Alph{section}}
\setcounter{section}{0}

\section{Code usage}

The code for this study is available online at \url{https://github.com/williamgilpin/SHREC}. Tutorials and docmentation for usage of the method are available in the repository, and the code uses the same conventions as the widely-used \texttt{scikit-learn} library \cite{pedregosa2011scikit}. \texttt{SHREC} can be applied to an arbitrary time series dataset as:

\begin{verbatim}
# Load a time series dataset with
# shape (n_timepoints, n_series)
X, y_true = load_data()

# Initialize model with default hyperparameters
model = RecurrenceManifold()

# Estimate driving signal
y_recon = model.fit_predict(X)
\end{verbatim}

\section{Shared reconstruction algorithm implementation}
\label{algo}

\noindent\textbf{Calculating the driver adjacency graph.}  For each of the $N$ observation time series $\v{x}_k(t)$, $k \in \{1, 2, ..., N\}$, we lift the time series using time delays $\hat{\v{y}}_k(t)$ and then compute the pairwise distance graph $d^{(k)}_{ij} = || \hat{\v{y}}_k(i) - \hat{\v{y}}_k(j)||$. For the $i^{th}$ row of this graph, we find the set of $M$ smallest elements $\{d^{(k)}_{im}\}_{m=1}^M$, which correspond to the distances to the $M$ nearest neighbors of point $i$. The smallest member of this set determines the constant
\[
	\rho_i \equiv \min_m \{d^{(k)}_{im}\}.
\]
We next perform a numerical rootfinding calculation to find the value of $\sigma_i$ such that
\[
	\log_2 M = \sum_{m=1}^M e^{-\text{ReLU}\left(d^{(k)}_{im} - \rho_i\right) / \sigma_i},
\]
where $\text{ReLU}(x) \equiv \max(0, x)$. Having calculated $\rho_i$, $\sigma_i$, we transform the original distance matrix into an affinity matrix via the transformation
\[
	A_{ij}^{(k)} = e^{-\text{ReLU}\left(d^{(k)}_{ij} - \rho_i\right) / \sigma_i}.
\]
This matrix has been used in recent algorithms in topological data analysis \cite{mcinnes2018umap}, and it encodes key features of a fuzzy simplicial complex associated with the points in $\hat{\v{y}}_k(i)$. We define a consensus affinity matrix as the elementwise average across the simplicial complexes for all individual response variables, $A_{ij} \equiv \sum_{k=1}^K A_{ij}^{(k)}$.

After creating the recurrence graph $A$, we next seek to reconstruct the driving signal $\hat{z}(t)$ underlying this graph. This problem is related to the general distance geometry problem, which has previously been studied in numerous contexts \cite{donner2010recurrence}, including in structural biology in the context of molecular contact maps \cite{shi2021hi} and resonance spectra produced by nuclear magnetic resonance \cite{more1997global,liberti2014euclidean}. However, our particular problem bears several distinctions: Our matrix $A$ does not correspond to a literal distance matrix, but rather a consensus connectivity matrix across several signals; we seek a low-dimensional driving signal, which makes the problem simpler; additionally, each element of $A$ must reveal information about a unique driver state. Thus, our approach to inferring $\hat{z}(t)$ varies based on the expected type of driving signal.

\noindent\textbf{Continuous-time dynamics.} Given the aggregated adjacency matrix $A$, we calculate the graph Laplacian $L = D - A$, where $D_{ij} = \delta_{ij} \sum_k A_{ik}$. Our interpretation of $A$ as a transition matrix implies that the spectrum of the graph Laplacian will either reveal non-ergodicity in the form of a graph partition, or otherwise weak ergodicity in the form of long-time transients associated with slowly-relaxing states. Therefore either the leading eigenvector or the first non-constant eigenvector (known as the Fiedler eigenvector) contain information about our driving signal. A similar principle underlies widely-used Laplacian eigenmap algorithms (and related diffusion maps), allowing us to use existing optimized solvers to compute the reconstructed driver \cite{coifman2006diffusion,giannakis2012nonlinear}. We note that our general problem of inferring a signal from a topological ordering nodes in a graph is identical to pseudotime calculation, a method in systems biology that infers developmental trajectories from gene expression measurements of unsorted cell populations \cite{haghverdi2016diffusion,nadler2006diffusion}. 

\noindent\textbf{Discrete-time dynamics.}  A discrete-time driving signal contains only finite number of unique states $k \leq T$, and so we seek to assign a fixed label to every node in $A$. While the labelling scheme and number of labels is well-defined when the graph contains disjoint subgraphs, for real-world datasets spurious recurrences cause all nodes to become mutually reachable. Thus, we instead seek to optimize a set of $k$ discrete partitions, or cuts, to the aggregated connectivity graph that maximizes an objective function like cluster modularity or separability. Globally optimal $k$-cut of a graph is an NP-hard problem, and so computing the optimal partition requires checking all possible partitions---a combinatorial optimization task that becomes impractical even for short time series \cite{mezard2009information}. However, an approximate heuristic approach would be spectral clustering, in which we apply our continuous time method, and then discretize the reconstructed signal into $k$ bins \cite{mehta2019high,girvan2002community}. However, the driver dynamics on this symbol space will not necessarily represent a faithful coarse-graining of the underlying dynamics, undermining the accuracy of the resulting reconstruction. Instead, we opt to use Leiden clustering, a graph community detection heuristic that traverses and partitions the transition graph, and also determines the optimal value of $k$ and thus unique driver states \cite{traag2019louvain}. We take the progression of cluster labels visited by the response time series as an estimate of the discrete driver, and we note that all permutations of cluster labels are equivalent.

\subsection{Demonstration with known dynamical systems}

For the illustrative example featured in the main text, the driving signal corresponds to the dynamics of the Rossler equations,
\begin{align}
\dot{z_1} &= -z_2 - z_3 \notag\\
\dot{z_2} &= z_1 + a z_2 \notag\\
\dot{z_3} &= b + z_3 (z_1 - c)
\label{drive}
\end{align}
where we have set $a = 0.2$, $b = 0.2$, and $c = 5.7$.

The response signals correspond to the Lorenz equations, each of which is driven identically by $z_1(t)$,
\begin{align}
\dot{y_1}^{(k)} &= \sigma^{(k)} (y_2^{(k)} - y_1^{(k)}) + \kappa z_1\notag\\
\dot{y_2}^{(k)} &= y_1^{(k)} (\rho^{(k)} - y_3^{(k)}) - y_2^{(k)} \notag\\
\dot{y_3}^{(k)} &= y_1^{(k)} y_2^{(k)} - \beta^{(k)} y_3^{(k)}
\label{resp}
\end{align}
where the different response systems are indexed by $k \in \{1, 2, ..., N\}$ and the coupling strength $\kappa$ determines the relative strength of the driver forcing. We set $\kappa = 1/2$ for our experiments. The parameter values for each response system are drawn from the uniform distribution, centered around default values,
\begin{align}
\sigma^{(k)}  &\sim\,\mathcal{U}(\sigma_0, w) \notag\\
\rho^{(k)}  &\sim \,\mathcal{U}(\rho_0, w)\notag\\
\beta^{(k)}  &\sim \,\mathcal{U}(\beta_0, 10\,w) \notag
\end{align}
with parameter values $\sigma_0 = 10$, $\rho_0 = 28$, $\beta_0 = 8/3$ corresponding to chaotic dynamics. We set the width of the uniform distribution $\mathcal{U}$ to $w = 1$ in order to ensure diverse response dynamics.

In order to make the problem more difficult, and to highlight our method's robustness, we further corrupt our dataset with random nonlinear measurements. For each response system $k$, we define a random measurement function $p^{(k)} (x) = 1/\sqrt{2 \pi {s^{(k)}}^2} \exp( - (x - \mu^{(k)})^2)/(2 {s^{(k)}}^2))$, where $s^{(k)}  \sim \mathcal{U}(0, 5), \mu^{(k)} \sim \mathcal{U}(0, 10)$

Combining everything together, we create a response dataset from the highly-corrupted values of the first response variable $X \equiv \{p^{(k)}(y_1^{(k)} (t))\}_N$. From this time series ensemble, we create the estimate $\hat{z}_1(t)$ using our method.

\section{Benchmark Experiments}
\label{supp_benchmarks}

Here, we describe the benchmark models and our hyperparameter tuning procedure. We do not tune any hyperparameters for our own method, \texttt{SHREC}. We perform all hyperparameter tuning on the training dataset, and evaluate model performance on a held-out testing dataset. Importantly, all of our testing datasets correspond to distinct dynamical systems; depending on the experimental dataset, the testing data corresponds to a different animal, experimental trial, or regulatory network. All testing scripts and tabular results are available in our open-source repository.

\subsection{Experiment Design Overview}

For our experiments, we apply our method and others to a variety of datasets in which a known driving signal influences a large, diverse set of response variables. We train reconstruction models on \textit{only} the response time series, and then compare the resulting output to the true driving signal. We emphasize that our unsupervised algorithm and other benchmark methods do not receive the driver as inputs---the driver is withheld, and the inferred driver reconstruction produced by each algorithm is compared to the true driver only in order to validate method performance. In real-world applications, the true driver would be unavailable.

The datasets on which we test our method are chosen to represent diverse potential applications in which several observable variables are driven by an external signal:
(1) The turbulent motion of a set of diffusing tracer particles in a chaotic double gyre flow. The driver is long-wavelength sinusoidal forcing on the flow, while the response variables comprise the radial coordinates of individual particles---as would be observed via Doppler velocimetry. The training and validation datasets represent flows with different tracer particle initial positions and forcing amplitudes. There are $64$ response time series, each containing $3000$ timepoints.
(2) A fetal electrocardiogram time series. The driver is the maternal respiration signal, while the response variables comprise multi-point electrocardiogram recordings placed across the fetal body \cite{sulas2021non}. The training and validation datasets represent different patients in the original study. There are $24$ response time series, each containing $3000$ timepoints.
(3) Stochastic gene expression dynamics in a subset of the yeast metabolic gene regulatory network, as simulated by an existing experiment design package for cell biology \cite{schaffter2011genenetweaver}. The driver corresponds to the expression level of the transcriptional regulator YDR043C in response to a knockdown perturbation, and the response states are the expression levels of downstream genes in the network. There are $96$ response time series, each of which contains $4000$ timepoints.
(4) Species abundances of the $20$ most abundant phytoplankton in two lakes in Switzerland over a $20$ year period \cite{pomati2020interacting}. Due to sampling inconsistencies, over the period we study $28\%$ of the total data is missing, and so we downsample the data to a monthly average abundances and replace any missing time series by forward-filling, in which each missing value is replaced by the most recent non-missing value. The driver is the daily temperature of the lake. For the validation and testing datasets we use data from two different lakes in the original dataset. There are $20$ response time series, each of which contains $324$ timepoints.

We do not tune any hyperparameters for our own method during our experiments, but for the benchmark models with tunable hyperparameters we use a grid search and cross validation to find the best hyperparameters, and report performance on held-out test data. 

\subsection{Baseline models evaluated.} We consider a variety of classical and deep learning based time series models, with a particular emphasis on unsupervised methods used for time-dependent hidden state estimation. Whenever possible, we aim to use reference implementations with default hyperparameters used in published studies; however, we tune hyperparameters separately for each model and dataset combination, in order to ensure that each baseline are fairly evaluated.  Most time series models contain the equivalent of a time-lag order or timescale parameter, which determines the number of past timepoints that the model uses to calculate the hidden state at a given instant; we focus on tuning this hyperparameter for all models. For models that can learn latent time series of varying dimensionality, we also tune the dimensionality of this space, because smaller latent spaces usually regularize models.

\paragraph*{Mean.} We take a simple timepoint-wise average across all response time series, and use the result as our estimate of the driver.

\paragraph*{Slow Feature Analysis (SFA), Principal Components Analysis (PCA), Independent Components Analysis (ICA), Dynamical Components Analysis (DCA), and Canonical Correlation Analysis (CCA).} 

We apply both principal component analysis and independent components analysis across all response time series, treating each timepoint as an independent datapoint (thereby discarding temporal correlations). We also consider variants in which we lift the dimensionality of each input time series using time delays before computing the components, allowing temporal information to influence the representation. In order to apply canonical correlation analysis to time series, we create time-lagged copies of the signal in order to raise the effective dimensionality of each input time series, and we find canonical correlation coordinates relating past values to future values \cite{brunton2017chaos}. We perform canonical correlation between time series one time lag interval in the past, and one time lag interval in the future. We use \texttt{scikit-sfa}, a reference implementation of SFA by the original discoverers\cite{wiskott2002slow,blaschke2006relation}, and we do the same for DCA \cite{clark2019unsupervised}.

For all methods, we perform hyperparameter tuning over all combinations of time delay $\tau\in \{1, 5, 10, 20, 50, 100\}$ and number of latent dimensions $L \in \{1, 5, 10, 20, 50\}$. We score model performance on held-out testing data. Because we are reconstructing one-dimensional driving signals, we use the first or most "dominant" latent dimension, which is determined using a different heustic for distinct methods---for example, in PCA this is the mode that captures the most variance, while in SFA it is the slowest variance-capturing mode. Some methods, like PCA, learn the same representation regardless of the hyperparameter $L$, but for other methods smaller $L$ regularizes the model by forcing it to span fewer latent dimensions, materially affecting the learned representation. Thus, while we treat $L$ as a hyperparameter for all methods, it does not necessarily affect all models.

\paragraph*{Kalman Filter.} We test several variants of the Kalman filter, and we report results using the best model as determined by cross-validation on the training dataset. The standard Kalman filter is a linear Gaussian process, but we also consider the ensemble Kalman filter and unscented Kalman filter variants \cite{hamilton2019correcting,ott2004local,hamilton2016ensemble,hunt2004four}. Because neither our general problem, nor the datasets evaluated in our experimental task, have consistent spatiotemporal structure or known governing equations, we do not include the traditional extended Kalman filter among our baselines. We use an implementation of the Kalman Filter in the Python package \texttt{filterpy}, as well as the Python time series analysis library \texttt{sktime} \cite{loning2019sktime}. These implementations use expectation-minimization to solve for the parameters of the model. We perform hyperparameter tuning across different Kalman filter variants, as well over a varying number of latent state dimensions $L \in \{1, 5, 10, 20, 50\}$. We also augment the dimensionality of our response datasets using a variable number of time delays,  $\tau\in \{0, 1, 5, 10, 20, 50, 100\}$. After performing cross-validation to select the best model for each dataset, we evaluate performance on held-out test data.

\paragraph*{Fourier Principal Components Analysis (fPCA).} For each input time series, we generate $10^4$ random surrogate time series with matching power spectra using the amplitude-adjusted Fourier transform method. For the original time series, as well as all surrogates, we compute the power spectrum. The model has a single hyperparameter, $\alpha \in [0, 1]$, the significance threshold. If an amplitude in the training data power spectrum is greater than a fraction $\alpha$ of all surrogate spectra, then the corresponding frequency is deemed significant. A reconstructed driver signal is generated by producing a time series containing only the significant frequencies from the input time series, along with their corresponding phase shifts, and the resulting reconstructions are combined across all observation time series by computing their first principal component. We perform hyperparameter tuning over $10$ possible values of the significance threshold $\alpha$.

\paragraph*{Gaussian Process Factor Analysis (GPFA).} We use an implementation of GPFA from the \texttt{elephant} Python package \cite{yu2008gaussian,yegenoglu2015elephant}. We convert each response time series to a spike train by sampling an inhomogenous Poisson process with firing rate set by the time series values (which are rescaled to fall between $0$ and $1$). The sampling rate of the process is set to $1$ ms$^{-1}$, while for the GPFA model, the spike binning size is set to $1$ ms. When fitting the GPFA model, we perform hyperparameter tuning over all combinations trial count $N_{trial} \in \{5, 10, 20, 40, 800\}$ and number of latent dimensions $L \in \{1, 5, 10\}$. We score model performance on held-out testing data. At larger trial values, less information is lost when the deterministic response signal is converted into a spike train.

\paragraph*{Linear Gaussian State Space Model.} Related to the Kalman model and GPFA, this model assumes linear latent and observed dynamics, with noise drawn from a multivariate Gaussian process \cite{sarkka2013bayesian}. This model can be considered the equivalent of a hidden Markov model with continuous hidden states.

\paragraph*{Causal Dilated Autoencoder (cCNN).} We use an implementation of a two-layer causal dilated autoencoder, which has recently been shown to successfully extract meaningful featurizations for a downstream classification task even when trained without labels \cite{franceschi2019unsupervised,baiTCN2018}. This model performs one-dimensional trainable convolutions across time for the response time series, and then uses a series of dilations and additional convolutions to aggregate information across timescales. We modify this model's final layers to preserve the number of timepoints in the input dataset, and treat the latent space as a compressed representation of the input signal. We perform hyperparameter tuning over all combinations of learning rate $\eta \in \{10^{-1}, 10^{-2}, 10^{-3}\}$, convolution block width $W \in \{8, 16, 32, 64\}$, and number of latent dimensions $L \in \{1, 5, 10, 20, 50\}$. We score model performance on held-out testing data.

\paragraph*{Latent-Factor Analysis via Dynamical Systems (LFADS).} We use a reference PyTorch implementation of LFADS \cite{pandarinath2018inferring,sussillo2016lfads,prince2021}. We perform hyperparameter tuning over all combinations of learning rate $\eta \in \{10^{-1}, 10^{-2}, 10^{-3}\}$, batch size $B \in \{5, 20, 100\}$, and number of latent factors $L \in \{5, 10, 20, 50\}$; however, we leave many of the other hyperparameters for the method at their default values. We emphasize that the relative performance of LFADS may be even stronger than we observe, if other hyperparameters (such as architecture depth, or choice of optimizer) are tuned as well; however, here we focus on hyperparameters that have analogues in the other benchmark models, in order to ensure we are searching similar model spaces and performing a fair comparison. We score model performance on held-out testing data.

\paragraph*{Hirata et al. \& Nomura et al. Isomap.} This approach requires thresholding the individual distance matrix of each embedded time series in order to ensure a given sparsity (set to $10\%$ nonzero values in the original works) \cite{hirata2008reproduction,nomura2022superposed}. The binary distance matrices are combined into a consensus matrix using Boolean addition (if a given matrix element is nonzero in any matrix, it is nonzero in the final matrix). This consensus adjacency matrix is converted into a weighted distance matrix by using Dijkstra's algorithm to find the shortest path between each pair of nodes. The distance matrix is then embedded into a one-dimensional space using multidimensional scaling. The latter two steps correspond to the Isomap data embedding algorithm. We perform hyperparameter tuning over all sparsity thresholds in $\{10^{-3}, 10^{-2}, 10^{-1}, 0.2\}$.

\subsection{Accuracy Scores}

Because our method seeks dynamical couplings between the driver $z(t)$ and response $\mathbf{y}(t)$, we consider several measures of the accuracy of the reconstructed driving signal $\hat{z}(t)$ and the true driving signal  $z(t)$. For strongly-coupled systems, we consider several pointwise error measures such as the mean squared error (MSE), Spearman correlation (Spearman), the symmetric mean absolute percent error (sMAPE) the mean absolute scaled error (MASE), and the dynamic time warping distance (DTW) \cite{giorgino2009computing}. These different accuracy measures are used to assess forecast quality in various contexts \cite{hyndman2006another,gilpin2021chaos,godahewa2021monash,makridakis2022m5}. Because a driver influence can act after time delay, we compute sliding versions of these pointwise metrics across multiple time delays between the true driver $z(t)$ and its reconstruction $\hat{z}(t)$, and we report the maximum. However, we require that the time delay must not exceed $30$\% of the length of the original time series. in order to avoid spurious correlations due to edge effects.

However, the use of pointwise error metrics presumes that the reconstructed driving signal should instantaneously resemble the unknown driving signal, a condition that only occurs in a strongly-coupled regime in which the response subsystem dynamics are not only fully-synchronized to the driver, but also have a monotonic phase relationship at all times \cite{ye2015distinguishing,munch2022recent,sugihara2012detecting,monster2017causal}. While these conditions likely hold for many time series of interest, we are particularly interested in the more challenging case in which the subsystem dynamics are driven, but not fully determined by, the driver signal. In this case, the effect of the driver on the response subsystem may be filtered or nonlinearly suppressed depending on system state.

As a more general measure of reconstructed signal accuracy, we consider the more general problem of determining whether the reconstructed driver signal $\hat{z}(t)$ bears any functional relationship with the true driver $z(t)$. We employ two methods for calculating this relationship: the mutual information, and nonlinear Granger causality. The former can be efficiently estimated from two time series using differential entropy, which can be computed non-parametrically via density-based methods \cite{perez2008estimation,kraskov2004estimating,kozachenko1987sample,evans2008computationally,lombardi2016nonparametric}. For nonlinear Granger causality \cite{tank2018neural,bennett2022rethinking,rosol2022granger}, we first train a state-of-the art nonlinear forecasting model to predict $\hat{z}(t)$ based on its past values. We then train another model that seeks predict $\hat{z}(t)$ based on past values of both $z(t)$ and $\hat{z}(t)$. We compute the Wilcoxon score between the two forecasts in order to quantify the degree to which knowledge of $z(t)$ improves forecasting of $\hat{z}(t)$ This approach can be viewed as a nonlinear generalization of classical Granger causality, which bypasses the need to enforce stationarity (e.g. via differencing) by instead learning a forecast model that can accommodate multiple timescales, seasonality, or drift. For our nonlinear forecasting model, we use N-HiTS, a recently-developed artificial neural network that hierarchically decomposes time series a trainable basis \cite{challu2022n}. Earlier versions of this model  show state-of-the-art performance on a variety of forecasting benchmarks for both real-world time series and dynamical systems datasets \cite{gilpin2021chaos,oreshkin2019n}.

For our benchmark experiments, we compute all $7$ scoring metrics described here across all datasets and baseline models. We correlate their performance across different models and datasets in Figure \ref{metrics}. Because Spearman correlation and neural Granger casuality report values of similarity, rather than distance, we invert the values of these metrics before computing correlations. 

We find that the simpler time series measures like Spearman and dynamic time warping correlate strongly with more sophisticated methods like neural Granger causality on the datasets we consider in our experiments. We thus conclude that our testing datasets primarily fall within a strongly synchronized regime, and we defer to using conceptually-simpler Spearman correlation for all results reported in the main text.

\begin{figure}
{
\centering
\includegraphics[width=0.8\linewidth]{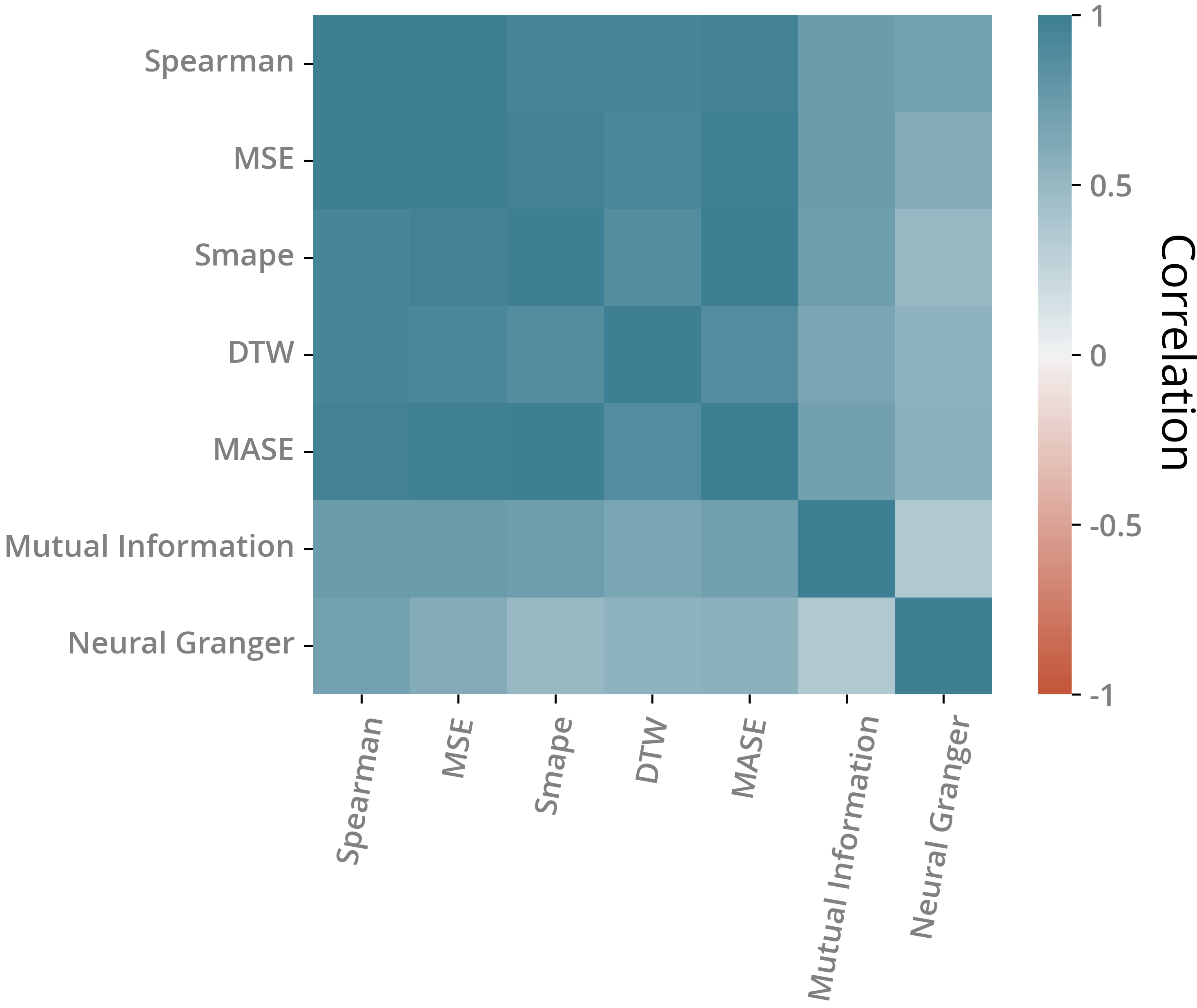}
\caption{
Spearman correlations among all metrics tested, aggregated across different datasets and benchmark reconstruction models.
}
\label{metrics}
}
\end{figure}

\begin{figure*}
{
\centering
\includegraphics[width=\linewidth]{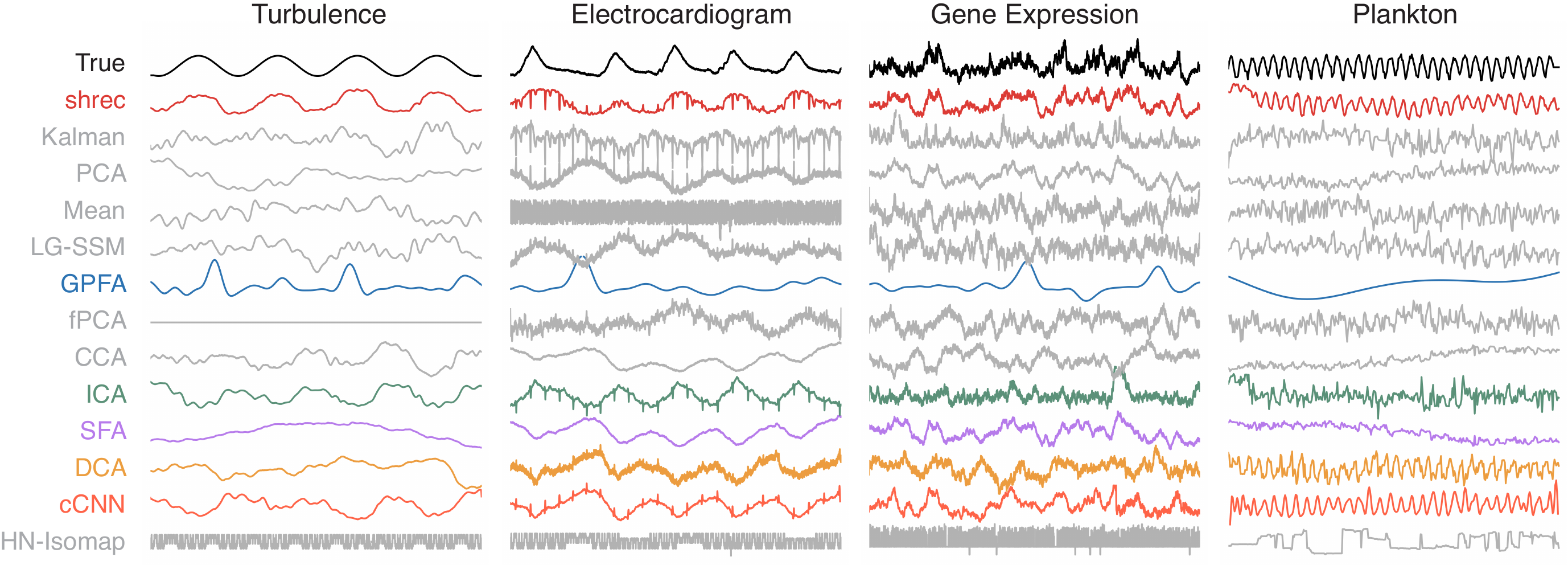}
\caption{
All reconstructed drivers. For methods accepting hyperparameters, the highest-scoring reconstruction (by Spearman correlation) is shown.
}
\label{metrics}
}
\end{figure*}

\section{Skew-product dynamical systems experiments}

Our dynamical systems database consists of $135$ crowdsourced chaotic dynamical systems, including well-known systems like the R\"ossler, Lorenz, and Chua equations, as well as domain-specific models from fields such as climatology, astrophysics, and neuroscience \cite{gilpin2021chaos,zhang2024zero}. To create random skew-product systems, we sample two distinct dynamical systems $\v{f}$ and $\v{g}$ from the database, and we label one as a driver and the other as a response.

The driver dynamics are generated autonomously $\dot{\v{z}}(t) = \v{f}(\v{z}(t))$. In the absence of driving, the response dynamics are given by $N$ systems indexed by $k$, $\dot{\v{y}}^{(k)}(t) = \v{g}^{(k)}(\v{y}^{(k)}(t))$. We drive these response dynamics by introducing a linear coupling scheme between the first variable of each response system, and the first variable of the driver, $\dot{y}^{(k)}_i(t) = g_i^{(k)}(\v{y}^{(k)}(t))+ \delta_{1i} \kappa_{f,g} z_1(t)$. The coupling constant $\kappa_{f,g}$ depends on the specific dynamical systems chosen to act as the driver and response, and it is intended to scale the driving to ensure that the relative amplitudes match across systems. We calculate $\kappa_{f,g}$ for each system pair by simulating the driver and response system separately (in the absence of coupling), and then calculating their average amplitudes and scaling the coupling appropriately, $\kappa_{f,g} = \kappa_0 \langle| y_1(t) |\rangle_t / \langle| z_1(t) |\rangle_t$. We set $\kappa_0$ equal to a fixed relative coupling strength for all skew product experiments.

We note that our illustrative example above, in which \eqref{drive} drives \eqref{resp}, constitutes one particular example of a skew product system constructed using our methods. For our large-scale experiments, we simulate dynamics for all $135 (135 - 1) = 18090$ possible unique, ordered pairs of dynamical systems; however, only $4952$ of these skew product systems are both numerically stable and exhibit chaotic dynamics (i.e., did not diverge or collapse to quiescence), and so we retain only these systems in our random skew product experiments. We note that weaker coupling, or more exhaustive searching of numerical integration tolerances or timesteps, might yield a greater number of stable skew product configurations.

For every skew product system, we apply our reconstruction method to the response dynamics, and measure accuracy as the timepoint-wise Spearman correlation with the driver time series (this is the same accuracy metric used in our benchmark experiments). Because every dynamical system in our database has been annotated with key mathematical properties (Lyapunov exponents, fractal dimension, entropy, etc), we can correlate the accuracy score with underlying Lyapunov exponent (a putative measure of chaoticity) separately for both the driver and response dynamical systems. Our Lyapunov exponent calculations are described in prior work \cite{gilpin2021chaos,gilpin2023large}; briefly, estimates are based on both a naive method based on initial exponential separation of trajectories over short and long periods (averaged over the attractor), and a more advanced method based on continuous orthonormalization of a bundle of vectors co-transported with the flow \cite{eckmann1986liapunov,abarbanel1991lyapunov}.

\section{Evaluating shared reconstruction algorithm properties across a period-doubling bifurcation}

While our benchmark experiments focus on continuous-time datasets due to their greater relevance to real-world problems, when examining our method's properties we include experiments with discrete time drivers, allowing us to probe the properties of our method across a period-doubling bifurcation leading to chaos. We follow the task originally used in Sauer's original paper describing the exact shared dynamics reconstruction algorithm \cite{sauer2004reconstruction}, which our approach seeks to generalize: an ensemble of logistic maps in the chaotic regime, but with random parameters, is driven by a logistic map in the periodic regime. We include the equivalent of Brownian stochastic forcing to the dynamics by adding to each iteration a small random number drawn from a normal distribution. We vary three parameters across our task as control parameters: the amplitude of the stochastic forcing, the strength of the coupling between the driver and the response series, and the number of response time series.

The driver time series is given by
$$
z_{t+1} = \lambda z_t (1 - z_t)
$$
This driver exhibits dynamics that are period $2$, $4$, and $8$ when $\lambda$ has values $3.4$, $3.5$, $3.5644072$, respectively, and it becomes fully chaotic above these values. The response time series are given by
$$
x_{t+1}^{(k)} = r^{(k)} x_t^{(k)} (1 - x_t^{(k)}) + \kappa z_t + \eta_t^{(k)}
$$
where the noise value is drawn from a normal distribution at each time step $\eta_t^{(k)} \sim \mathcal{N}(0, \sigma)$. The different response time series are labelled by $k = 1, 2, ..., N$, and each response system has distinct dynamical parameters $r^{(k)} \sim \mathcal{U}(r_0, \sigma_r)$. We set $r_0= 3.81$, a value well within the chaotic regime of the logistic map. We set $\sigma_r = 0.16$, $\sigma = 0.04$ and $\kappa = 0.5$ as default values throughout our experiments. For our percolation experiments, we separately vary as control parameters each of $\sigma$, $\kappa$, and $N$, while holding the other parameters fixed.

\subsection{Accuracy score for discrete time driving signals}

For the percolation experiments, we evaluate the accuracy of our method using the adjusted Rand score, a measure of pointwise accuracy that is invariant to permutations in labels. For example, if the period-two driver exhibits a sequence of states of the form $z_t = ABAB ...$, then a perfect adjusted Rand score of $1$ can be either of the two possible reconstruction sequences $\hat{z}_t = ABAB...$ or $\hat{z}_t = BABA...$. We note that, unlike the unadjusted Rand score, the adjusted Rand score is scaled so that random assignment of labels to timepoints will return a score of zero, rather than the nonzero pointwise accuracy resulting from random assignment. This correction ensures that scores generated for datasets with different period drivers remain comparable.

\subsection{Naive expectation for scaling of the accuracy score}

In the case of a discrete-time, periodic driver, we can generate a naive hypothesis for the scaling of our method's accuracy with the number or length of response time series. We assume that overall accuracy is primarily determined by the total number of recurrence events observed, $N_r$, which we naively expect to be directly proportional to the product $N T$, where $N$ is the number of response time series and $T$ is their average length. However, if the driver has more unique states (indicating a higher period $\tau$), then more observations are needed in order to resolve each of the distinct states. We thus assume that the overall accuracy should have a functional dependence on the number of recurrences divided by the driver period $N_r / \tau = N T / \tau$.

Each recurrence event reveals information about a finite number of potential driver states, but some recurrences will reveal redundant information about driver states that have already been fully-resolved. As $N T / \tau$ increases, we expect the number of novel discovered driver states to gradually saturate. This distribution is given by the survival function of a Bernoulli trial, or equivalent the mean of a beta distribution (which serves as the conjugate distribution to the Poisson distribution). 
\[
	\text{Acc} = \text{Acc}_\text{max}\left(1 - \exp(- \beta \sqrt{N T / \tau})\right)
\]
where $\text{Acc}_\text{max}$ and $\beta$ are constants that depend on the particular dynamical system. For the results shown in the main text, we fit a single value of $\beta$ to all three driver periods $\tau$ tested, but otherwise do not perform separate fits for different $\tau$.

\subsection{Percolation calculation}

In order to measure the structural properties of our model during fitting, we extract and measure the aggregated time series adjacency matrix $A$ before finding the driver with the diffusion. In order to understand the structure of the graph defined by this matrix, we compute the order parameter $T_{LCC} / T \in [0, 1]$, where $T$ is the total number of nodes and $T_{LCC}$ is the number of nodes comprising the largest connected component. As this value approaches one, it indicates the onset of percolation in finite-size undirected networks \cite{hebert2019smeared},

\section{Comparing the unstable periodic orbits of the reconstructed driver and original driver}

\subsection{Detecting Unstable Periodic Orbits from data}

A set of $M$ candidate unstable periodic orbits $\{ O_j \}$ of varying lengths $O_j \in \mathbb{R}^{\ell \times D}$ are detected directly from a given time series $X \in \mathbb{R}^{T \times D}$ by searching for close recurrences in the pairwise Euclidean distance matrix: $d(X, X) < \epsilon$, where $d(X, X)  \in \mathbb{R}^{T \times T}$\cite{lathrop1989characterization}. Each detected orbit $O_j$ is aligned in time to the original time series by maximizing its cross-correlation, and the resulting phase shift is used to calculate a new, periodic single-orbit time series $\hat{O}_j$ with the same number of timepoints as the original time series $\hat{O}_j \in \mathbb{R}^{T \times D}$ by repeating the orbit multiple times. The singular value decomposition of the set of aligned orbits $\{ O_j \} \in \mathbb{R}^{M \times T \times D}$ is then calculated in order to find a set of $m \leq M$ unique orbits, where the truncation point $m$ is determined by the drop-off rate in the sorted singular values \cite{brunton2022data}. The original time series $X$ is then projected onto the reduced orbit basis, in order to determine the relative amount of time each of the $m$ leading orbits are shadowed by the dynamics. We note that all operations in the singular value decomposition and basis set projections are computed as double contractions over the time and dimensionality axes, $X : \hat{O}_j \in \mathbb{R}$.

For the R\"ossler attractor dataset we use as an illustrative example throughout the main text, the spanning orbit sets comprises the first $m=3$ orbits. We note that a longer-duration or higher-time-resolution sample from the dynamical system might yield a spanning set with different cardinality.

\subsection{Comparing orbits and distance matrices}

Given two time series of equal lengths, such as a dataset time series $X \in \mathbb{R}^{T \times D}$ and an aligned orbit time series $\hat{O}_j \in \mathbb{R}^{T \times D}$, we compare the time series directly by contracting over both indices $X : \hat{O}_j \in \mathbb{R}$. The set of aligned orbits $\{\hat{O}_j \}$ can therefore be converted into the equivalent of a normal basis by rescaling $\hat{O}_j \leftarrow \hat{O}_j  / (\hat{O}_j  : \hat{O}_j)$.

However, when comparing two distance matrices, $d(X, X) \in \mathbb{R}^{T \times T}$ and $d(\hat{O}_j , \hat{O}_j ) \in \mathbb{R}^{T \times T}$, we are primarily interested in whether the matrices have similar overall structure (i.e., shared recurrences and long-term patterns). Therefore, instead of using mean-squared error, we use the element-wise Pearson correlation to compare $d(X, X)$ to $d(\hat{O}_j , \hat{O}_j)$ across various conditions. 

In the main text, we compare the distance matrix of the reconstructed signal and the first three orbits, after aligning the latters' phases to the reconstructed driver. As a control parameter, we vary the number of response time series $N$, and thus the final quality of the reconstruction.

\section{Fuzzy simplicial complexes are a robust alternative to recurrence plots.}
\label{app_simplex_recur}
\begin{figure}
{
\centering
    \includegraphics[width=\linewidth]{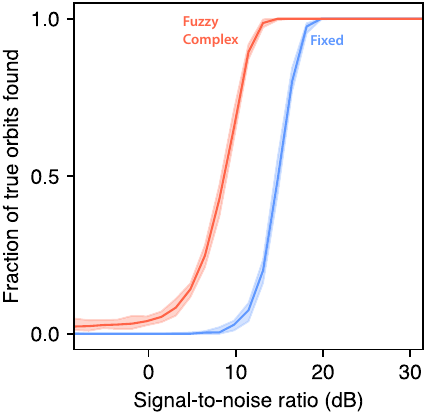}
\caption{
\textbf{Robustness of recurrence detection to noise.} 
The fraction of correctly-detected recurrence events relative to the amount of random noise. Error bars are 95\% confidence intervals, bootstrapped over random time series subsegments and noise initializations. Blue traces correspond to recurrences found using a fixed distance threshold $\epsilon$, while red traces correspond to recurrences detected using the adaptive fuzzy approach described in the main text.
}
\label{fuzzy}
}
\end{figure}

We generate a $10^4$-timepoint sample of dynamics from the Lorenz equations. The sampling timescale is chosen such that the dynamics span $40$ Lyapunov times. From this time series, we set a fixed recurrence threshold $\epsilon$, and record the set $\mathcal{B}_0$ of all timepoints of recurrences, $\mathcal{B}_0 \equiv \{(i,j)\}$ such that $\| \v{x}(t_i) - \v{x}(t_j) \| \leq \epsilon$.

We next add random Gaussian noise to each timepoint, $\tilde{\v{x}}(t_i) \equiv \v{x}(t_i) + \eta_i$, $\eta_i \sim \mathcal{N}(0, \eta)$. We recompute the set of recurrences, $\mathcal{B}_\eta \equiv \{(i,j)\}$ such that $\| \tilde{\v{x}}(t_i) - \tilde{\v{x}}(t_j) \| \leq \epsilon$. We compute an accuracy score as the true positive rate $\text{Acc}(\eta) \equiv \abs{\mathcal{B}_\eta \cap \mathcal{B}_0} / \abs{\mathcal{B}_\eta}$. We find that other metrics, such as the F1-score, give similar results. We convert the added noise into a dimensionless signal-to-noise ratio, 
\[
    \text{SNR}(\eta) \equiv 10 \log_{10}\left( \frac{x_\text{rms}}{\eta} \right),
\]
where $x_\text{rms} \equiv \sqrt{\langle \v{x}(t_i) \cdot \v{x}(t_i) \rangle_i}$.

We find that fuzzy simplicial complexes are able to maintain accurate detection of recurrences even when the SNR decreases by a factor of $10$ (Fig. \ref{fuzzy}). Moreover, fuzzy simplicial complexes do not require the recurrence threshold $\epsilon$ to be chosen in advance.

\section{Alternative community detection methods for driver reconstruction.}
\label{app_community}

\begin{figure}
{
\centering
    \includegraphics[width=\linewidth]{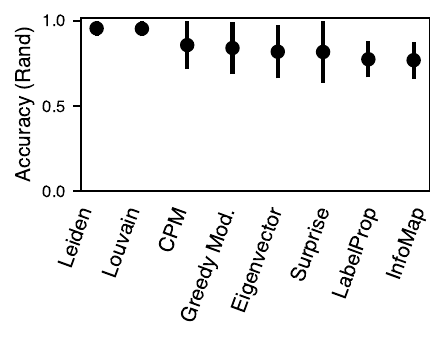}
\caption{
\textbf{Comparison of community detection methods for discrete-time drivers.} An ensemble of $200$ logistic maps in the chaotic regime is driven by a single shared driver in the period-4 regime. Noise is added to the response dynamics with a signal-to-noise ratio of $20$ dB. Accuracy is scored using the adjusted Rand Index compared to the $4$ possible true driver states. Bars show median and interquartile range across 60 replicates with random initial conditions and algorithm seeds.
}
\label{community}
}
\end{figure}

In the discrete-time version of our algorithm, we make the particular choice to use Leiden community detection in order to assign driver states to equivalence classes in the presence of noise. We opt for this algorithm due to recent work showing that it balances accuracy and cost, particularly for large networks \cite{traag2019louvain}. To evaluate the importance of this choice, in Figure \ref{community} we repeat the experiment design from Section \ref{sec_percolation}. We choose a single value for the noise and driving, corresponding to a case in Figure \ref{properties} where nearly exact reconstruction was possible. We compare Leiden to Louvain clustering, a similar algorithm, as well as a variety of other community detection methods \cite{girvan2002community,newman2006finding,blondel2008fast,raghavan2007near,reichardt2006statistical,clauset2004finding,aldecoa2011deciphering}. We observe that all methods perform comparably on this task, including eigenvector-based clustering, which resembles the approach that we use for continuous-time driver reconstruction in the other sections. Several of the methods, including InfoMap and LabelProp, are not optimized for this particular task (for which distinct driver states induce well-separated communities). We thus interpret this performance comparison narrowly in the context of causal driver reconstruction. However, our results overall do not appear to be particularly sensitive to the choice of clustering heuristic.

\section{Detecting higher-order interactions in gene regulatory networks}
\label{simplex_app}

\begin{figure}
{
\centering
\includegraphics[width=\linewidth]{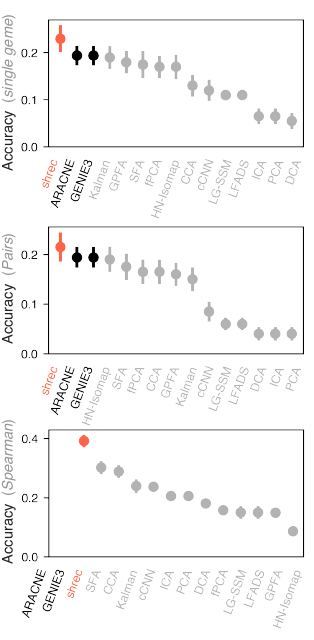}
\caption{
\textbf{Reconstruction of higher-order terms in a gene regulatory network is robust to the choice of accuracy metric.} The single gene accuracy compares the reconstructed driver to the dynamics of each gene time series, in order to detect the two highest scoring genes. These genes are then compared to the true driving genes. The combination gene accuracy compares the reconstructed driver to the products of all possible pairs of time series, and compares the two genes in the highest scoring pair to the true driver genes. The Spearman accuracy measures the timepoint-wise correlation between the reconstructed driver dynamics and true combination driver. ARACNE and GENIE3 directly infer the interaction network among time series, and then select the top two highest-scoring genes by outdegree. For this reason, these methods do not produce time series and thus cannot be scored by Spearman correlation, and they also yield the same results for the two accuracy scores.
}
\label{simplex_metrics}
}
\end{figure}

Following prior work \cite{marbach2010revealing,schaffter2011genenetweaver,mendes2003artificial}, we implement a mechanistic model of gene expression that separately models abundances of mRNA and protein products associated with each gene. In the absence of interactions, we denote the dynamics of an isolated node as
\[
	\dot{\v{X}}_k(t) = \v{f}(\v{X}_k(t) ).
\]
where $\v{X}_k(t) \in \mathbb{R}^2$ for a model that includes separate dynamical variables for the mRNA and protein abundances associated with each gene. Hereafter, we drop the vector notation and use $X_k(t) \in \mathbb{R}$ to denote the dynamical variable associated with interactions among genes over experimentally-relevant timescales.

A traditional regulatory network model couples the dynamics of individual genes together additively through the traceless coupling matrix $A$,
\[
	\dot{X}_k(t) = f(X_k(t) ) + \sum_{\ell \neq k} A_{k \ell} {X}_\ell(t).
\]
Network inference techniques therefore seek an estimate of the interaction matrix $\hat{A}$ given a set of measurements, including direct observations of the relevant dynamical variables $\{ X_k(t) \}_N$.
In order to capture higher-order multiplicative interactions, we introduce an additional term to each gene's dynamics, which depends on the product of the expression levels of two other genes,
\[
	\dot{X}_k(t) = f(X_k(t) ) + \sum_{\ell \neq k} A_{k \ell} {X}_\ell(t) +  \sum_{\substack{\ell \neq k, m \neq k \\ m \neq \ell}}\!\! B_{k \ell m} X_m(t) X_\ell(t)
\]
Importantly, methods that seek to directly infer $A$ cannot necessarily capture the interaction structure of $B$. However, in some cases network reconstruction methods may assign entries to their estimate of the interaction matrix $\hat{A}$ that result from interactions in $B$. For example, if $B_{123} \neq 0$, then a method might find $\hat{A}_{12}, \hat{A}_{13} \neq 0$. 

For our experiments, we impose exactly one driving combination among the set of $N$ genes. In this case, each row of $B_{:,\ell m}$ contains only one nonzero term $B_{k \ell m} = B_{k} \delta_{\ell \alpha}\delta_{m \beta}$. The dynamics then simplify,
\begin{equation}
	\dot{X}_k(t) = f(X_k(t) ) + \sum_{\ell \neq k} A_{k \ell} X_\ell(t) +  B_{k} X_\alpha(t) X_\beta(t).
\label{simplex_dyn}
\end{equation}
Identification of the causal driving therefore requires correctly identifying the two indices $\alpha,\beta \in 1,2,...,N$. Time-resolved driver reconstruction methods, including SHREC and the methods benchmarked in the previous section, produce a driver estimate $\hat{y}(t)$, which can be directly compared to the true multiplicative driver $y(t) = X_\alpha(t) X_\beta(t)$. Regulatory network reconstruction methods, however, instead directly return estimates $\hat{\alpha}, \hat{\beta}$

We generate a set of $150$ random networks by randomly sampling connected subnetworks of $N=12$ genes from a subset of the yeast metabolic network, which was also used in the DREAM4 challenge \cite{marbach2010revealing,schaffter2011genenetweaver}. For each network, we randomly select two nodes to act as multiplicative drivers, and we then add a random forcing term to the remaining $N-2$ genes' dynamics, following \eqref{simplex_dyn}. For each of the $150$ networks, we generate $N=12$ time series of length $T=100$ timepoints.

We set aside $50$ networks for validation, and we use these to find the median best hyperparameter values for the various causal reconstruction methods, using the same methods and procedures described in Appendix \ref{supp_benchmarks}. We also use two methods that directly estimate the time series interaction network, GENIE3 and ARACNE \cite{huynh2010inferring,margolin2006aracne}. We do not tune any hyperparameters for our method, SHREC. 

We test our methods on the remaining $100$ random networks with higher-order interactions. We evaluate the effectiveness of different methods in uncovering the higher-order driving using three metrics: 

The \textbf{Spearman correlation} directly compares the reconstructed driver $\hat{\mathbf{y}}(t)$ with the true driver $y(t) \equiv X_\alpha(t) X_\beta(t)$.
\[
	\text{Acc.} = \rho(\hat{y}(t), X_{\alpha}(t) X_{\beta}(t))
\]

 The \textbf{Detection accuracy} finds the two largest of $N$ pairwise correlations between the reconstruction $\hat{y}(t)$ and any individual time series $X_i(t)$, and estimates $\hat{\alpha}, \hat{\beta} \in 1, 2, ... N$ as the indices of the two most correlated expression time series. These indices are then compared to the true indices $\alpha, \beta$ to generate an accuracy score, which can be $0$, $0.5$, or $1$ depending on the number of driver genes correctly detected,
\[
	\hat{\alpha}, \hat{\beta} = \underset{a \neq b}{\arg\max} \, \left\{ \rho(X_{a}(t), \hat{y}(t)), \rho(X_{b}(t), \hat{y}(t)) \right\}
\]
\begin{equation}
\text{Acc.} = 
\begin{cases} 
1 & \text{if } \{\hat{\alpha}, \hat{\beta}\} = \{\alpha, \beta\} \\
0.5 & \text{if } \{\hat{\alpha}, \hat{\beta}\} \cap \{\alpha, \beta\} \neq \emptyset \text{ and } \{\hat{\alpha}, \hat{\beta}\} \neq \{\alpha, \beta\} \\
0 & \text{if } \{\hat{\alpha}, \hat{\beta}\} \cap \{\alpha, \beta\} = \emptyset
\end{cases}
\label{acc}
\end{equation}

The \textbf{Combination accuracy} instead tests all $N(N-1)$ pairwise products $X_a(t) X_b(t)$ among the $N$ observed time series, and returns the indices of the single most correlated pair $\hat{\alpha}, \hat{\beta}$,
\[
	\hat{\alpha}, \hat{\beta} = \arg\max_{a,b} \rho\left( X_a(t) X_b(t), \hat{y}(t) \right)
\]
The highest scoring single pair $\hat{\alpha}, \hat{\beta} \in 1, 2, ... N$ is then compared to the true indices $\alpha, \beta$ via \eqref{acc} to generate a score, which can also be $0$, $0.5$, or $1$ depending on the number of driver genes correctly detected. For each baseline method evaluated, we average its score across all $100$ networks in the testing set.

We find that all three metrics largely agree (Fig. \ref{simplex_metrics}), particularly among the higher-performing methods, with the shared reconstruction algorithm consistently detecting the multiplicative driving genes more effectively than other methods. The specialized regulatory network inference methods, ARACNE and GENIE3, consistently perform well, suggesting that these methods, while originally developed for traditional network inference, also identify signatures of higher-order interactions. 

Our particular problem setting represents a small regulatory network with a single higher order driver. We emphasize that our experiments are not intended to show that our method, SHREC, should be used instead of existing network inference methods in biology \cite{hecker2009gene}. Rather, this example highlights that our physics-based method performs surprisingly well on a network inference problem with biological motivation. Our results therefore underscore the strong inductive biases of recurrence-based methods, allowing generalization to diverse problem settings.

\newpage
\bibliography{shrec_cites} 
\bibliographystyle{naturemag}

\end{document}